\documentclass[letterpaper]{article} 
\usepackage{aaai2026}  
\usepackage{times}  
\usepackage{helvet}  
\usepackage{courier}  
\usepackage[hyphens]{url}  
\usepackage{graphicx} 
\usepackage{natbib}  
\usepackage{caption} 
\usepackage{algorithm}
\usepackage{algorithmic}

\usepackage{newfloat}
\usepackage{listings}
\DeclareCaptionStyle{ruled}{labelfont=normalfont,labelsep=colon,strut=off} 
\floatstyle{ruled}
\newfloat{listing}{tb}{lst}{}
\floatname{listing}{Listing}
\usepackage{amsmath}
\usepackage{csquotes}
\usepackage{subcaption}
\usepackage{longtable}
\usepackage{xcolor}
\usepackage{makecell}
\usepackage{verbatim}
\usepackage{listings}
\usepackage{graphicx}
\usepackage{tikz}
\usepackage{multirow}
\usepackage{listings}
\usepackage{xcolor}
\usepackage{xcolor}
\usepackage{verbatim}
\usepackage{booktabs}
\usepackage{caption}
\title{Seeing Like an AI: How LLMs Apply (and Misapply) Wikipedia Neutrality Norms}

\author{
    Joshua Ashkinaze\textsuperscript{\rm 1},
    Ruijia Guan\textsuperscript{\rm 1},
    Laura Kurek\textsuperscript{\rm 1},
    Eytan Adar\textsuperscript{\rm 1},
    Ceren Budak\textsuperscript{\rm 1},
    Eric Gilbert\textsuperscript{\rm 1}
}

\affiliations{
    \textsuperscript{\rm 1}University of Michigan\\
    jashkina@umich.edu, rjguan@umich.edu, lkurek@umich.edu, eadar@umich.edu, cbudak@umich.edu, eegg@umich.edu
}

\usepackage{bibentry}

\begin{document}

\maketitle

\begin{abstract}
Large language models (LLMs) are trained on broad corpora and then used in communities with specialized norms and rules. But can LLMs apply community rules well enough to follow these norms? We evaluate LLMs' capacity to detect (Task 1) and correct (Task 2) biased Wikipedia edits according to Wikipedia's Neutral Point of View (NPOV) policy. LLMs struggled with bias detection, achieving only 64\% accuracy on a balanced dataset. Models exhibited contrasting biases (some under- and others over-predicted bias), suggesting distinct priors about neutrality. LLMs performed better at bias correction, removing 79\% of words removed by Wikipedia editors. However, LLMs made additional changes beyond Wikipedia editors' simpler neutralizations, resulting in high-recall but low-precision editing. Interestingly, crowdworkers rated AI rewrites as more neutral (70\%) and fluent-sounding (61\%) than Wikipedia-editor rewrites. Qualitative analysis found that LLMs sometimes applied NPOV more comprehensively than Wikipedia editors but often made extraneous non-NPOV-related changes (e.g., grammar). LLMs may apply rules in ways that resonate with the public, but diverge from community experts. While potentially effective for generation, LLMs may reduce editor agency and increase moderation workload (e.g., verifying additions). Even when rules are easy to articulate, having LLMs apply them \textit{like community members} may still be difficult.
\end{abstract}

%
\begin{links}
    \link{Supplementary Materials}{https://osf.io/6h3cp}
\end{links}

\section{Introduction}

Large language models (LLMs) are trained on large, broad corpora but then used within smaller communities that have their own norms. To steer models towards specific norms and values, there is a growing trend of stating high-level rules as prompts. For example, constitutional AI~\cite{bai_constitutional_2022} involves providing models with explicit rules that guide their behavior. But is providing high-level rules sufficient to steer models toward community norms? 

The challenge of going from high-level rules---like Wikipedia's Neutral Point of View (NPOV) policy\footnote{Avoid stating opinions as facts; Avoid stating seriously contested assertions as facts; Avoid stating facts as opinions; Prefer nonjudgmental language; Indicate the relative prominence of opposing views. (See \url{https://en.wikipedia.org/wiki/Wikipedia:Neutral_point_of_view}.)}---to specific cases mirrors earlier debates in human-AI interaction and beyond. For instance, Lucy Suchman's 1987 Plans and Situated Actions framework~\cite{suchman_plans_1987} contrasted predetermined procedures derived from universal principles (``plans'') with context-dependent actions based on concrete circumstances (``situated actions''). She argued that AI systems execute plans, while humans perform situated actions. This distinction relates to James Scott's~\cite{scott_seeing_1998} concept of ``seeing like a state''---the idea that large-scale plans of centralized authorities often break down when faced with complex local realities. The tension between universal plans and situated local actions is relevant to deploying general-purpose LLMs in communities like Wikipedia. Many decisions involve contextual judgments~\cite{hansen_wikipedia_2009, swarts_collaborative_2009}. Is providing high-level rules to trillion-parameter models \textit{also} insufficient for navigating particular NPOV cases?

We provide the first comprehensive evaluation of how well general-purpose LLMs can be steered to apply NPOV. The broader NPOV policy encompasses various assessments (e.g., covering viewpoints in proportion to their prominence in reliable sources). Here, we focus on one subtask: debiasing language. Specifically, we evaluate LLMs' ability to: (1) detect edits that include non-neutral language and (2) generate NPOV-compliant edits from biased ones. Evaluating LLMs on Wikipedia's NPOV policy is an interesting test case of model abilities for three reasons. First, it provides a practical use case for determining whether these models can be steered---with no task-specific fine-tuning---toward following community policies or guidelines. Second, detecting and generating non-neutral language requires nuance that goes beyond literal meaning. This task probes how well LLMs can apply natural language pragmatics and mirror the nuanced decisions of community members. Third, Wikipedia's NPOV has a unique tension: while clearly-articulated and well-documented in theory, it is complex in practice~\cite{reagle_is_2007, swarts_collaborative_2009}. Adjudicating neutrality is nuanced. To what extent can LLMs, absent fine-tuning, apply clearly articulated but nuanced community policies?

\subsection{Research Questions}
Concretely, our research questions are:
\begin{description}
    \item \textbf{RQ1:} Can LLMs detect NPOV versus non-NPOV edits (and what factors affect performance)?
    \item \textbf{RQ2:} How do LLM neutralizations differ from neutralizations by Wikipedia editors?
    \item \textbf{RQ3:} How do crowdworkers rate LLM neutralizations compared to Wikipedia-editor neutralizations on key factors such as neutrality and fluency?  
\end{description}

\subsection{Findings}

To test whether LLMs could apply Wikipedia's NPOV policy, we conducted detection and generation experiments. LLM decisions and edits were compared against an existing corpus of NPOV edits by Wikipedia editors, also known as Wikipedians. In the detection experiment, we used ChatGPT 3.5, Mistral-Medium, and GPT-4 to classify whether edits violated NPOV. In the generation experiments, we used GPT-4 (using both Zero-Shot and Constitutional AI approaches) to neutralize NPOV-violating edits. We used computational metrics to compare the LLM-neutralized edits to the Wikipedian-neutralized edits. In the second generation experiment, humans gave masked evaluations of LLM and Wikipedian rewrites. We also conducted a qualitative analysis of LLM versus Wikipedian rewrites.

\begin{enumerate}

    \item \textbf{Large language models largely failed at neutrality detection.} Across models, prompts, and optimization strategies, performance was low (64\% accuracy for the best prompt compared to chance accuracy of 50\%). Analysis of LLM rationales and errors suggested that LLMs relied (sometimes to a fault) on simple heuristics such as highly subjective adjectives. 
    
    \item \textbf{Models exhibited contrasting biases that persisted across prompts.} ChatGPT 3.5 over-predicted edits as biased, and Mistral-Medium did the opposite. GPT-4 was balanced. Pretrained LLMs may internalize distinct priors about what constitutes neutrality.

   \item \textbf{LLMs applied rules in different ways than expert editors.} Computational experiments showed that in the generative phase LLMs typically removed words that Wikipedia editors did, but also made many additional changes. In other words, LLM editors are high-recall but low-precision. And while Wikipedians make more removals than additions, LLM editors do the opposite. 

  \item \textbf{Crowdworkers preferred LLM neutralizations to Wikipedia-editor neutralizations.}  
  We conducted human experiments to understand if these generations are \textit{preferred} to Wikipedian neutralizations. We find that crowdworkers prefer AI edits over human edits on both fluency and bias reduction. 

\item \textbf{Qualitative analysis showed LLMs are `NPOV+'.} To reconcile LLMs' low detection performance with high preference evaluations, we conducted a qualitative analysis. Despite being instructed to make minimal (NPOV) changes, LLMs made many grammatical and stylistic edits that may have influenced participants' judgments. Such edits could (1) increase labor costs, as moderators may need to check for AI hallucinations, and (2) reduce editor agency due to extensive rewrites. But in other cases, AI models arguably applied NPOV more faithfully than human editors (as judged by the authors).
\end{enumerate}

\section{Related Work}

We position our work in the context of Wikipedia neutrality research (\S \ref{wiki_background}), automated moderation of Wikipedia (\S \ref{wiki_auto}), and broader research on the role of general-purpose LLMs in community moderation (\S \ref{pre_trained_auto_mod}). Wikipedia's NPOV policy, though clearly defined, is often debated in practice (\S \ref{wiki_background}), making it a compelling test case for LLMs. While prior work tested \textit{task-specific} models on neutralizing NPOV violations, we focused on the effectiveness of general-purpose LLMs. This task is timely given two trends. First, recent work has explored whether pre-trained LLMs can apply community policies (\S \ref{pre_trained_auto_mod}). Second, Wikipedia has historically used automated moderation\footnote{\url{https://en.wikipedia.org/wiki/Wikipedia:WikiTrust}}, but there are contentious internal debates about the use of ChatGPT~\cite{harrison_should_2023}. Our study can inform these debates. More broadly, our work speaks to the opportunities and limitations of using general-purpose LLMs to uphold nuanced community norms.

\subsection{Wikipedia's NPOV Policy}

\label{wiki_background}
Wikipedia is one of the most visited websites in the world\footnote{\url{https://www.semrush.com/website/top/}}. It is a widely-regarded success in peer production. To support a crowd-sourced encyclopedia editable by anyone, Wikipedia developed many policies, processes, and norms~\cite{butler_dont_2008}. 
``Neutral Point of View'' (NPOV) is one of the most central. For example, NPOV is listed first among the three core content policies~\cite{wikipedia_core_2023} and Jimmy Wales (a co-founder of Wikipedia) said that ``the biggest and the most important thing is our neutral point of view policy.''~\cite{wales_jimmy_2006}. NPOV is crucial for Wikipedia's content quality--- defending against disinformation, low-quality information, and delineating fringe points of view~\cite{steinsson_rule_2024, mcdowell_it_2020}. 

Because of NPOV's importance, Wikipedia goes to great lengths to codify what a ``neutral point of view'' means. Specifically, NPOV is described by a set of principles that are expounded and clarified in FAQs, tutorials, and examples. Editors are trained in NPOV before they can edit. Articles violating NPOV are flagged as ``NPOV Disputed'' so that an editor can bring them into compliance. As a community norm, NPOV is clearly defined and often invoked in day-to-day community activity. 

However, despite its clear articulation, NPOV is complicated in practice. Reagle claims that ``in the Wikipedia culture, the notion of `neutrality' is not understood so much as an end result, but as a process''~\cite{reagle_is_2007}. Wikipedia articles often go through many rewrites, with debates occurring as to what constitutes neutrality. For example, over three years, the Wikipedia entry for `clean coal technology' had 39 distinct facts and 142 different rewrites of these facts~\cite{swarts_collaborative_2009}. Norm-consistent Wikipedia content is often the result of conflict, coordination, and deliberation---more complex than lone editors applying rules in isolation~\cite{kittur_he_2007}. Past work~\cite{kittur_harnessing_2008} estimated that 40\% of Wikipedia edits are dedicated to coordination between Wikipedians (e.g., achieving consensus).

Wikipedia's NPOV policy is an interesting test case of general-purpose LLM abilities because of a tension: It is both clear \textit{and} nuanced. On the one hand, NPOV is clearly defined and documented. In theory, augmenting an LLM with the NPOV guidelines text should enable the LLM to apply these guidelines effectively. But even though NPOV is clear in theory, it is complex in practice. To what extent can general-purpose LLMs---absent task-specific training---be steered to follow nuanced guidelines such as NPOV?

\subsection{Automated Approaches for NPOV}
\label{wiki_auto}

Prior studies explored neutralizing NPOV-violating edits, conducting experiments on edits where the bias was attributable to only one word~\cite{recasens_linguistic_2013, pryzant_automatically_2020}. Recasens et al. found that both humans and their system struggled to identify biased words (accuracy: 37\% and 34\%). Later, Pryzant et al. expanded on this work, collecting NPOV-violating edits and their corrections to form the Wikipedia Neutrality Corpus (WNC). Pryzant et al. trained a seq-to-seq model that achieved 46\% generation accuracy (as the proportion of generations that matched Wikipedia-editor rewrites). We leverage this dataset in our work.

Our study departs from these studies in two ways. First, we conduct experiments on a more representative set of NPOV-violating edits (i.e., not just the one-word subset of the WNC), which is arguably a harder task. Several other studies have experimented with multi-word neutralization as well, using cross-domain adaptive pretraining~\cite{madanagopal_improving_2022}, reinforced sequence training~\cite{madanagopal_reinforced_2023}, and cyclic bias neutralization with auxiliary guidance~\cite{madanagopal_bias_2023}. Second, we test whether general-purpose models, absent any task-specific training, can perform this task. Prior studies show that applying NPOV in a way that matches how expert editors behave is difficult for both laypeople \textit{and} older, task-specific NLP systems. In our work, we ask: How well can new generations of general-purpose LLMs neutralize text?

Wikipedia already uses automated moderation tools\footnote{\url{https://en.wikipedia.org/wiki/Wikipedia:Automated\_moderation}}~\cite{halfaker_ores_2020, rawat_automatic_2019}. More recently, Wikipedia editors have debated the utility of ChatGPT for generating Wikipedia edits~\cite{harrison_should_2023}. Our results can directly inform the benefits and risks of using general-purpose LLMs on Wikipedia.

\subsection{Pre-Trained LLMs for Moderation}
\label{pre_trained_auto_mod}

Given how much content there is to moderate, platforms often employ automatic content moderation tools. General-purpose LLMs show remarkable zero-shot~\cite{gilardi_chatgpt_2023} and few-shot~\cite{wang_generalizing_2020} capabilities even without fine-tuning (i.e., training the model on task-specific data). Fine-tuning requires resources (compute, instances) that some communities may not have. This has motivated research on the use of off-the-shelf LLMs to perform online community moderation. 

\citet{kolla_llm-mod_2024} tested whether LLM moderators can detect rule-violating posts across nine subreddits when the rules of these subreddits are provided to the LLM. Their system achieved a high true-negative rate (92\%) but a low true-positive rate (43\%). A similar study found that ChatGPT 3.5 was 64\% accurate in predicting subreddit moderation decisions~\cite{kumar_watch_2024}. \citet{cao_toxicity_2024} tested how well LLMs could detect rule violations of r/AskHistorians, with precision and recall showing variance depending on the specific rule. 

Our study complements and extends these studies by testing whether general-purpose LLMs can effectively apply Wikipedia NPOV. NPOV is both intrinsically important (being a central norm to one of the most visited websites) and theoretically interesting (since it is highly documented \textit{and} nuanced). Furthermore, in addition to detection, we focus on how well LLMs can \textit{generate} content consistent with a community's rules. As LLM usage grows, understanding the implications for \textit{generation} within communities is crucial. We conduct both computational and human subject experiments to measure generative quality. Because LLMs are explicitly trained for natural language generation, they may be more effective at generation than detection.


\begin{table*}[t]
\centering
\begin{tabular}{lllllll}
\toprule
Prompt & \multicolumn{2}{l}{Minimal} & \multicolumn{2}{l}{NPOV} & \multicolumn{2}{l}{NPOVScoped} \\
Examples &      ZeroShot &       FewShot &      ZeroShot &       FewShot &      ZeroShot &       FewShot \\
Model         &               &               &               &               &               &               \\
\midrule
ChatGPT 3.5   &  0.57 (0.029) &  0.56 (0.029) &  0.53 (0.029) &  0.55 (0.029) &  0.53 (0.029) &  0.57 (0.029) \\
MistralMedium &  0.57 (0.029) &  0.61 (0.028) &  0.60 (0.029) &  0.57 (0.029) &  0.59 (0.029) &  0.58 (0.029) \\
GPT-4         &  0.59 (0.028) &  0.61 (0.028) &  0.63 (0.028) &  0.59 (0.028) &  0.61 (0.028) &  0.60 (0.028) \\
\bottomrule
\end{tabular}
\caption{Accuracy of models and prompts. SEs are in parentheses.}
\label{acc_table}
\end{table*}

\section{LLM Bias Detection}
To understand whether general-purpose LLMs can classify edits as neutral or biased, we conducted classification experiments on a balanced sample of NPOV-violating edits and NPOV-compliant rewrites. We varied both how neutrality was defined (Factor 1) and whether examples were shown (Factor 2). We find that across models and prompts, models largely failed to distinguish between neutral and biased edits. Models exhibited contrasting failure modes, and model explanations suggest they appeared to rely on heuristics such as the presence of a highly subjective adjective.  

\subsection{Dataset}

We use the Wikipedia Neutrality Corpus (WNC)~\cite{pryzant_automatically_2020} (MIT License) for our experiments. The WNC contains edit pairs: a biased edit (flagged by an editor for violating NPOV) and a neutral edit (the rewrite to ensure NPOV compliance). The WNC focuses on violations of a subset of NPOV rules---biased language (framing, epistemological, and demographic bias). This subset is useful for testing LLMs since biased language does not require external information. For detection, we use edits tagged by topic (Appendix Table \ref{tab:topic_counts} for topic counts). 

The WNC was collected by crawling 423,823 Wikipedia revisions between 2004 and 2019~\cite{pryzant_automatically_2020}. The original data was first filtered for revisions, where editors provided an NPOV-related justification. The authors optimized the precision of bias-related changes by using a series of rule-based logic that, for example, dropped revisions where only non-literary elements or grammar were changed and excluded revisions below a minimal and above a maximal number of characters. From these filtered revisions, \citet{pryzant_automatically_2020} then identified specific sets of (biased, neutralized) sentence pairs using BLEU~\cite{papineni_bleu_2002}. These sets of (biased, neutralized) edit pairs comprise the WNC. While there are other datasets related to NPOV violations, we chose the WNC over more heavily curated datasets (e.g., WIKIBIAS~\cite{zhong_wikibias_2021}) as it contains naturally occurring NPOV-flagged edits directly from Wikipedia without extensive manual filtering, reflecting how LLMs would encounter content in real-world applications.

\subsection{Experiment Setup}
We conducted a multi-model prompt experiment ($N$ = 5,358 annotations) where the task was to classify if a given Wikipedia edit was biased or neutral. Our two experimental factors were \textbf{(Factor 1)} the definition given to an LLM on what constitutes neutrality, and \textbf{(Factor 2)} whether or not we provided examples (i.e., few-shot or zero-shot). The rationale for these factors is described later in this section. Appendix \ref{appendix_prompts} has all the prompts we used in this paper.

\begin{itemize}
    \item \textbf{Factor 1: Definitions}
    \begin{itemize}
        \item Minimal definition (\textbf{Minimal})
        \item Wikipedia's Neutral Point of View definition (\textbf{NPOV})
        \item Wikipedia's definition focused on neutral language (\textbf{NPOV-Scoped})
    \end{itemize}
    \item \textbf{Factor 2: Examples}
    \begin{itemize}
        \item No examples: Zero-shot (\textbf{ZeroShot})
        \item With examples: 10-shot examples from the same topic (\textbf{10-Shot})
    \end{itemize}
\end{itemize}

We employed three models: \textit{gpt-4-0125-preview, gpt-3.5-turbo}, and \textit{mistral-medium-latest}. We refer to these as \textit{GPT-4, ChatGPT 3.5}, and \textit{Mistral-Medium} from here on. We chose these on the basis that---at the time of this study---the first is a state-of-the-art LLM, the second is commonly used (largely due to speed and cost), and the third is from a popular open-source developer. For each of the 18 (model $\times$ definition $\times$ example) combinations, we classify a balanced (between biased edits and neutralized rewrites) sample of 300 edits stratified by topic, yielding 5,400 attempted classifications. For each classification, we instruct the model to return the correct label (`neutral' or `biased'), a rationale, and the policy violated (if the prompt contains policies). Of 5,400 attempted annotations, 5,358 yielded compliant annotations (i.e., one of the two classes). We classified this number of edits due to API costs associated with expensive state-of-the-art models such as GPT-4. Confidence intervals for quantities of interest are narrow.

\subsubsection{Factor 1: Definitions Provided}
\label{factor_1}
The \textbf{minimal} prompt relies on LLMs' learned knowledge only, providing no Wikipedia-specific definitions. This condition reveals how well general-purpose LLMs apply NPOV without additional context. The \textbf{NPOV} prompt provides LLMs with Wikipedia's verbatim NPOV guidelines. Comparing the performance between the \textbf{minimal} and \textbf{NPOV} conditions estimates how much simple prompting can align general-purpose notions of neutrality with community-specific norms. Since the WNC focuses on a subset of NPOV violations related to non-neutral language, we created a third prompt condition: \textbf{NPOV-Scoped}. This prompt gives LLMs Wikipedia's NPOV guidelines and additional language from Pryzant et al. on the specific types of non-neutral language (framing, epistemological, and demographic bias) in the dataset. Comparing \textbf{NPOV-Scoped} to \textbf{NPOV} reveals how much guideline specificity improves classification accuracy.

\subsubsection{Factor 2: Examples Provided}
\label{factor_2}
Few-shot learning~\cite{wang_generalizing_2020} can improve model performance. We test whether augmenting the model with editors' prior decisions increases accuracy. In the few-shot condition, we select a random sample of 10 edits and their labels from the same topic, as neutrality norms may differ by topic due to factors like similar editors or content.

\begin{figure*}[h]
    \centering

        \centering
        \includegraphics[width=1\linewidth]{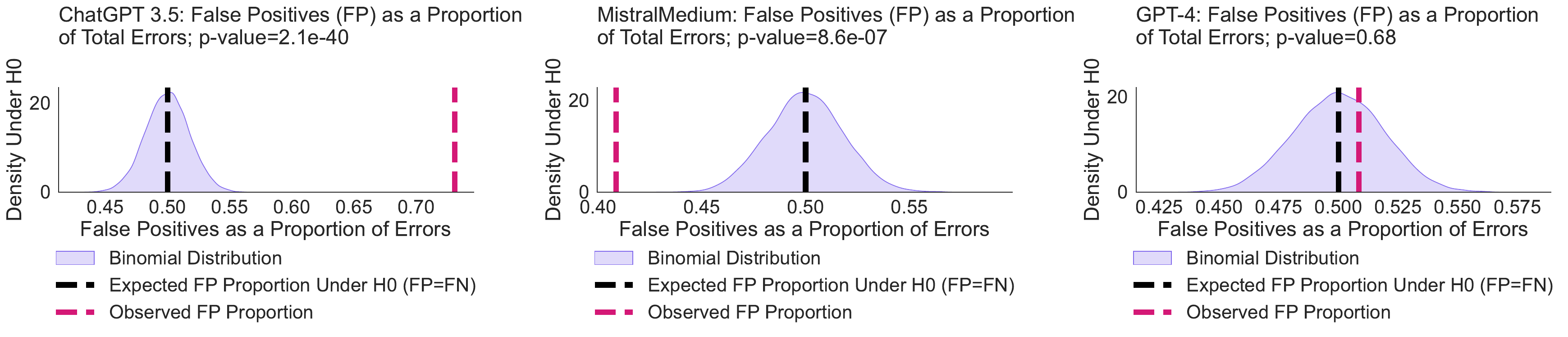}
        \caption{Model error imbalance tests. We simulated a null hypothesis of evenly divided errors, and then compared each model's observed error balance to this distribution (Appendix \ref{binom_test} for details).}
        \label{fig:binom}

    \label{fig:combined}
\end{figure*}

\subsection{Results}

\paragraph{Main Results}

\label{overall}

All models performed poorly, and prompts made little difference (Table \ref{acc_table}). Across all conditions, the accuracy was 0.58. Averaging across models, ChatGPT 3.5 performed the worst with an accuracy of 0.55, and GPT-4 performed the best at 0.61. The top combination was GPT-4 with a zero-shot NPOV prompt (0.63). For each of the models, there were no statistically significant differences in accuracy between prompt conditions, based on p-values from two-tailed permutation tests. Models were more accurate for biased edits (0.63, 95\% CI = [0.61, 0.65]) than neutral edits (0.53, 95\% CI = [0.51, 0.55]), two-tailed permutation $p<0.001$. We also found that examples tempered ChatGPT 3.5's predictions: ChatGPT 3.5 predicted 83\% of edits to be biased under zero-shot prompts but only 59\% of edits to be biased under few-shot prompts, two-tailed permutation $p<0.001$. See Appendix Table \ref{logitacc} for regressions. 

\paragraph{Additional Experiments}
For robustness, we conducted further experiments using more advanced self-optimization and reasoning techniques (e.g., auto-prompting and CoT) (Appendix \ref{self_opt}) as well as a bias-identification approach (Appendix \ref{bias_based}). These experiments did not yield improvements above our main prompts, which are the subject of further analysis in this section.

\paragraph{Model-Level Analysis}
\label{model}
We examined the error distribution of different models. Not all models failed the same way (Figure \ref{fig:binom}, Appendix Figure \ref{fig:confusion} for confusion matrices). ChatGPT 3.5 was far more likely than other models to predict that edits were biased when they were neutral. Mistral-Medium, on the other hand, erred in the opposite direction, over-predicting neutral edits. GPT-4 was balanced. We assessed the statistical significance of these patterns with two-tailed binomial tests (Appendix \ref{binom_test}; Figure \ref{fig:binom} for more details), rejecting a null hypothesis of balanced errors for ChatGPT 3.5 and Mistral-Medium but not GPT-4. Because prompts and specific edits were held constant, these differences in model predictions arise from the model itself. Our results suggest that prompting may not be enough. If models have idiosyncrasies, these may persist through prompts.

\paragraph{Edit-Level Analysis}

We calculated the edit-level probability of correct classification (PCC) to identify which edits are easier to detect. PCC is the probability of a model correctly classifying an edit across all models and prompts. PCCs were bimodal (Appendix Figure \ref{ridge}), indicating that edits tend to be either easy or hard to classify. 

We conducted a qualitative analysis of `easy' vs. `hard' biased edits (i.e., top- or bottom-PCC quartile). Broadly, top-PCC biased edits tend to have some highly subjective word that alerts models that these edits are biased. Here is one example (emphasis added, Appendix \ref{edit_level_app} for more examples): ``\textit{one of the central characters of the novel, akili kuwale, provides a \textbf{brilliant} demonstration of this change and its implications, together with \textbf{excellent} characterization.}'' Bottom-PCC biased edits did not have such words.

\paragraph{Explanation-Level Analysis}
\label{explanation}

We prompted models to provide explanations along with predictions. We conducted a TF-IDF logistic regression to understand what specific words in a model explanation correlate with accuracy. We used an 80-20 train-test split over \{Explanation, Is Accurate\} tuples and 5-fold cross-validated grid search to tune hyperparameters. The best model achieved 0.8 accuracy, 0.8 precision, and 0.79 recall. Referencing highly subjective words (Appendix Figure \ref{fig:logit_explanation} for top features), such as `sadly', was associated with accuracy. Occasionally, the model flagged edits as biased when they contained a subjective-sounding word, even when the edit itself was not biased. As an example, models misclassified the following edit as biased (emphasis added): 

\begin{displayquote}
the advertisement called obama a \textbf{hypocrite} for not supporting armed guards in schools while noting that the children of the us president receive special protection by armed agents of the us secret service.
\end{displayquote}

In this edit, the writer is referencing the advertisement---but LLMs flagged the edit as biased even though it was factually recalling what an \textit{ad} said. GPT-4's explanation:

\begin{displayquote}
 The edit uses loaded language by calling Obama a `hypocrite' which is an opinion stated as a fact, and also uses judgmental language which violates the neutral point of view policy.
\end{displayquote}

We discuss more examples in Appendix \ref{qual_exp}. These dynamics suggest that LLMs relied (sometimes to a fault) on simple heuristics, like the presence of a highly subjective adjective. However, LLMs did effectively \textit{identify} the subjective-sounding part of the text in many of these false positive cases. This suggests that general-purpose LLMs may be able to neutralize biased text---but possibly at the cost of precision. This is what we find in the next section. 

\section{LLM Neutrality Generation: Computational Evaluation}

Section 3 showed that LLMs struggle to detect neutral vs. non-neutral edits. However, once a \textit{human editor} flags an edit as NPOV-violating, perhaps LLMs can neutralize the edit. Here, we evaluate the LLMs' ability to generate neutral edits from biased ones. To do this, we compared LLM neutralizations of biased edits to Wikipedian neutralizations of the same biased edits. 

\begin{table*}[ht]
\centering
\small 
\renewcommand{\arraystretch}{1.3} 
\begin{tabular}{@{}lll@{}} 
\toprule
\textbf{Metric} & \textbf{Practical Interpretation} & \textbf{Operationalization} \\
\midrule
Precision & Does AI change only what is needed? & $\text{TP} / (\text{TP} + \text{FP})$ \\
Recall & Does AI `catch' words that human editors flag? & $\text{TP} / (\text{TP} + \text{FN})$ \\
F1 & A balance (harmonic mean) of precision and recall & $2 \cdot \frac{\text{Precision} \cdot \text{Recall}}{\text{Precision} + \text{Recall}}$ \\
Similarity & How similar are AI changes to human ones? & $|A \cap B| / |A \cup B|$ \\

BLEU Score & How similar is AI-generated text to human text?   & Common machine-translation metric \cite{papineni_bleu_2002} \\

Non Disjoint & Are AI changes at least somewhat accurate?  & $ 1 \text{ if } |A \cap B| > 0; \, 0 \text{ otherwise}$ \\
\bottomrule
\end{tabular}
\caption{Description of natural language generation accuracy metrics. These metrics compare AI neutralizations to neutralizations from Wikipedia editors, treating the latter as the gold standard. To distinguish: `Similarity' compares \textit{changes} from the original edit, while `BLEU Score' compares the \textit{final edits'} similarity.}
\label{tab:metrics}
\end{table*}

\begin{figure*}[t]
\centering

\begin{minipage}{\textwidth}
\centering
\includegraphics[width=0.86\linewidth]{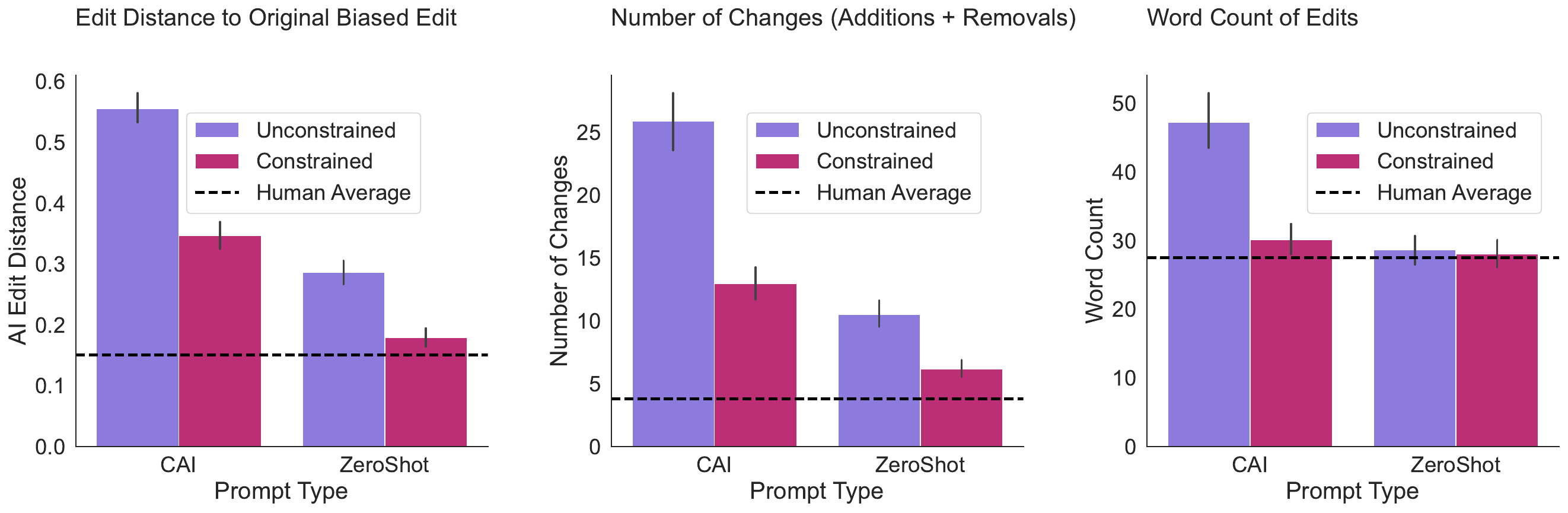}
\caption{Automated metrics (95\% CIs) for edit intensity. The horizontal line is the average for Wikipedia editor rewrites.}
\label{fig:combined_auto_met_simp}
\end{minipage}

\end{figure*}

\subsection{Experiment Setup}
\paragraph{LLM Generation}
We use our highest-performing model (GPT-4) and the NPOV-Scoped\footnote{We used this prompt since GPT-4 performance did not significantly differ between any prompt condition, but the dataset contains violations of neutral language, in particular.} prompt to neutralize biased Wikipedia edits. Specifically, we instructed the model to revise an NPOV-violating edit, varying two factors. We varied \textbf{(Factor 1)} whether we conducted this generation using zero-shot reasoning or Constitutional AI (CAI). In the latter, the model first critiqued why an edit violated NPOV and then used that critique to revise the edit. Since initial results suggested LLMs changed more words than human editors, we experimented with instructions to only edit what was necessary \textbf{(Factor 2)}. We conducted this generation on 200 edits, varying both prompt type (CAI or Zero-Shot) and whether to add constraining instructions (Yes or No), yielding 800 generations. For all conditions, we set the temperature to zero to further reduce extraneous changes. 

\paragraph{Measures}
\label{gen_measures}

We evaluated both the \textit{intensity} and \textit{accuracy} of AI changes. For intensity, we computed (1) the normalized edit distance between the AI-neutralized and original NPOV-violating edit, (2) the word count of AI edits, and (3) the number of changed words (excluding stopwords) in each AI edit. We compared these metrics to analogous metrics for human-modified Wikipedia edits to determine if models make more changes than human editors.

For accuracy, we compared AI edits to Wikipedia editors' changes, which we treated as the gold standard. Our diff-based approach is similar to that of Pryzant et al. We computed diffs between original biased edits ($w_{i}^{\text{Bias}}$), human-modified edits ($w_{i}^{\text{ModH}}$), and AI-modified edits ($w_{i}^{\text{ModAI}}$). After preprocessing (lowercasing, removing punctuation and stopwords), we defined sets of words removed by AI ($A$) and humans ($B$). From these, we calculated true positives ($|A \cap B|$), false positives ($|A - B|$), and false negatives ($|B - A|$). We computed metrics from these quantities (Table \ref{tab:metrics}) and BLEU scores using human edits as a reference text. Undefined precision was treated as 0, undefined recall as missing, and we used BLEU score smoothing from~\citet{lin_orange_2004}. As robustness checks (Appendix Figure \ref{fig:multiverse}), we re-analyzed the data under 32 different combinations of analytical choices (e.g., removing stopwords) to ensure our findings were not artifacts of analytical decisions. Analyses yielded results similar to those in Table \ref{ai_removal}.

\subsection{Results}

\textbf{\textit{Finding 1: AI changes more than humans.}} On average, AI (pooling across all AI conditions) changed more than humans (Figure \ref{fig:combined_auto_met_simp}). AI edit distance ($M$ = 0.34, $SD$ = 0.21) was larger than human edit distance ($M$ = 0.15, $SD$ = 0.13), $t$(799) = 24.06, $p<0.001$, Cohen's $d_z$ = 0.85. Considering both additions and removals, AI edits had more changes ($M$ = 13.88, $SD$ = 12.91) than human edits ($M$ = 3.79, $SD$ = 4.01), $t$(799) = 22.10, $p<0.001$, $d_z$ = 0.78. When using CAI, AI edits were longer ($M$ = 33.53, $SD$ = 21.19) than human edits ($M$ = 27.53, $SD$ = 14.98), $t$(799) = 10.87, $p<0.001$, $d_z$ = 0.38. See Appendix Table \ref{regautomet} for regressions. Within AI conditions, CAI makes more changes than zero-shot, and constraining instructions are more effective in reducing changes for CAI than for zero-shot.

\begin{figure}[h!]
    \centering
    \includegraphics[width=1\linewidth]{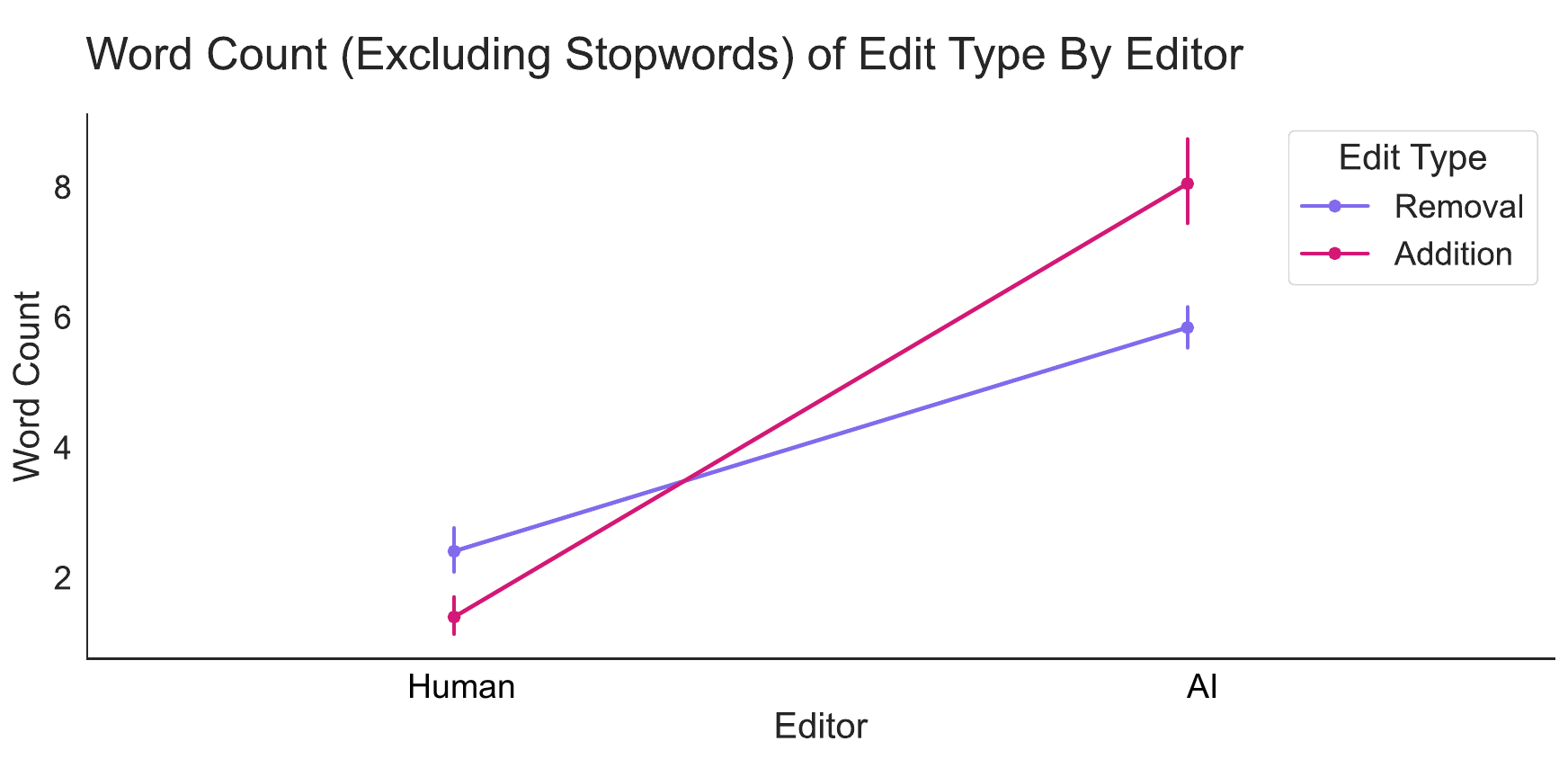}

    \caption{AI neutralizes edits via adding words and humans neutralize edits via removing words. Error bars are 95\% CIs.}
    \label{fig:edit_pref}
\end{figure}

\begin{table*}
\centering
\begin{tabular}{llllrlll}
\toprule
Prompt Type & Is Constrained & Precision & Recall &   F1 & Similarity & Non Disjoint & BLEU \\
\midrule
        CAI &              N &      0.29 &   0.89 & 0.44 &       0.27 &         0.90 & 0.23 \\
        CAI &              Y &      0.36 &   0.82 & 0.50 &       0.33 &         0.86 & 0.44 \\
   ZeroShot &              N &      0.37 &   0.77 & 0.50 &       0.34 &         0.82 & 0.48 \\
   ZeroShot &              Y &      0.47 &   0.67 & 0.56 &       0.41 &         0.76 & 0.64 \\
\bottomrule
\end{tabular}
\caption{Comparing AI edits to edits by Wikipedia editors. In general, AI editors had higher recall than precision.}
\label{ai_removal}
\end{table*}

\textbf{\textit{Finding 2: Humans are more likely to remove while AI is more likely to add.}} We find different editing patterns for AI vs human editors (Figure \ref{fig:edit_pref}). AI additions ($M$ = 8.04, $SD$ = 9.33) were higher than removals ($M$ = 5.83, $SD$ = 4.61), $t$(799) = 8.86, $p<0.001$, $d_z$ = 0.31. Meanwhile, human removals ($M$ = 2.40, $SD$ = 2.31) were higher than additions  ($M$ = 1.39, $SD$ = 2.14), $t$(199) = 7.34, $p<0.001$, $d_z$ = 0.52. Within AI conditions, condition-level dynamics of addition/removal (Appendix Figure \ref{fig:combined_change_plots_horizontal}, Appendix Table \ref{regautoaddrem}) are similar to Finding 1.

\textbf{\textit{Finding 3: AI changes are high recall but low precision.}}
Across AI conditions, recall (0.79; 95\% CI = [0.76-0.81]) was twice as high as precision (0.37; 95\% CI = [0.35-0.40]). While AI editors remove content that human editors remove, AI editors also make many other changes that human editors do not. See Table \ref{ai_removal} for condition-level breakdowns and see Appendix Table \ref{regautometacc} for regressions. We also computed precision and recall using a semantic approach (Appendix \ref{BertScore}) and found that the same pattern directionally held.

\section{LLM Neutrality Generation: Human Evaluation}
Computational experiments found that AI editors make more changes than Wikipedia editors. But which edits do humans prefer? We conducted an experiment to address this question\footnote{This experiment was approved by our university's IRB.}. Participants were shown one edit that initially violated NPOV and two sets of (masked) annotated revisions---one from a Wikipedia editor, and one from an AI condition (Zero-Shot and CAI, both with constraints). Participants rated rewrites on: (1) bias reduction; (2) adding/removing information from the original (apart from NPOV changes), (3) fluency. 

\subsection{Experiment Setup}
\paragraph{Participants}
We recruited 147 participants through Prolific, a crowdsourcing platform. Each participant completed 10 trials. Participants received \$3.33. Pilot tests indicated the experiment would take 11 minutes. This sample size was larger than that required by Orme's rule of thumb~\cite{orme_sample_1998} for discrete choice experiments. We restricted our study to users who: were over 18 years old, lived in the United States, completed more than 100 Prolific tasks, and had a 98\%+ approval rating. An ideal sample would be a large group of Wikipedians, but this is quite difficult to assemble. Instead, we turn to a sample approximating Wikipedia's readership---which can inform some central questions of this work but has other limitations. These limitations are discussed further in subsequent sections. 

\paragraph{Procedure}
After giving their informed consent, participants read an introduction to neutral language on Wikipedia. Then participants completed a training module where they rated the neutrality of 3 edits [-2, 2] and received feedback and an explanation after each of these examples. Next, participants continued to the main task. For each trial, participants saw one original edit and two visually annotated revisions---one from a Wikipedia editor, and one from an AI condition (in a randomized order and masked). 44 sets of the form \{Original Biased Edit, Human-Modified Edit, Zero-Shot Modified Edit, CAI-Modified Edit\} were used for the experiment.\footnote{Before the experiment, we standardized the capitalization of edits and fixed the spacing of punctuation marks in WNC edits. This was informed by the feedback from pilot studies.} First, participants selected which rewrite most increased neutrality. Second, participants answered whether each rewrite (A) removed information from the original edit (yes/no) and (B) added information to the original edit (yes/no), \textit{excluding} changes that increased the neutrality of the edit. Third, participants picked which rewrite was most fluent. After 10 trials, participants were fully debriefed and asked to guess what proportion of time other participants said the AI edit increased neutrality more than the human edit (to see if AI outperformed lay expectations). See the Supplementary Material file for experiment screenshots with task wording. 


\begin{figure}[t!]
    \centering
    \begin{subfigure}[t]{0.48\textwidth}
        \centering
        \includegraphics[width=\linewidth]{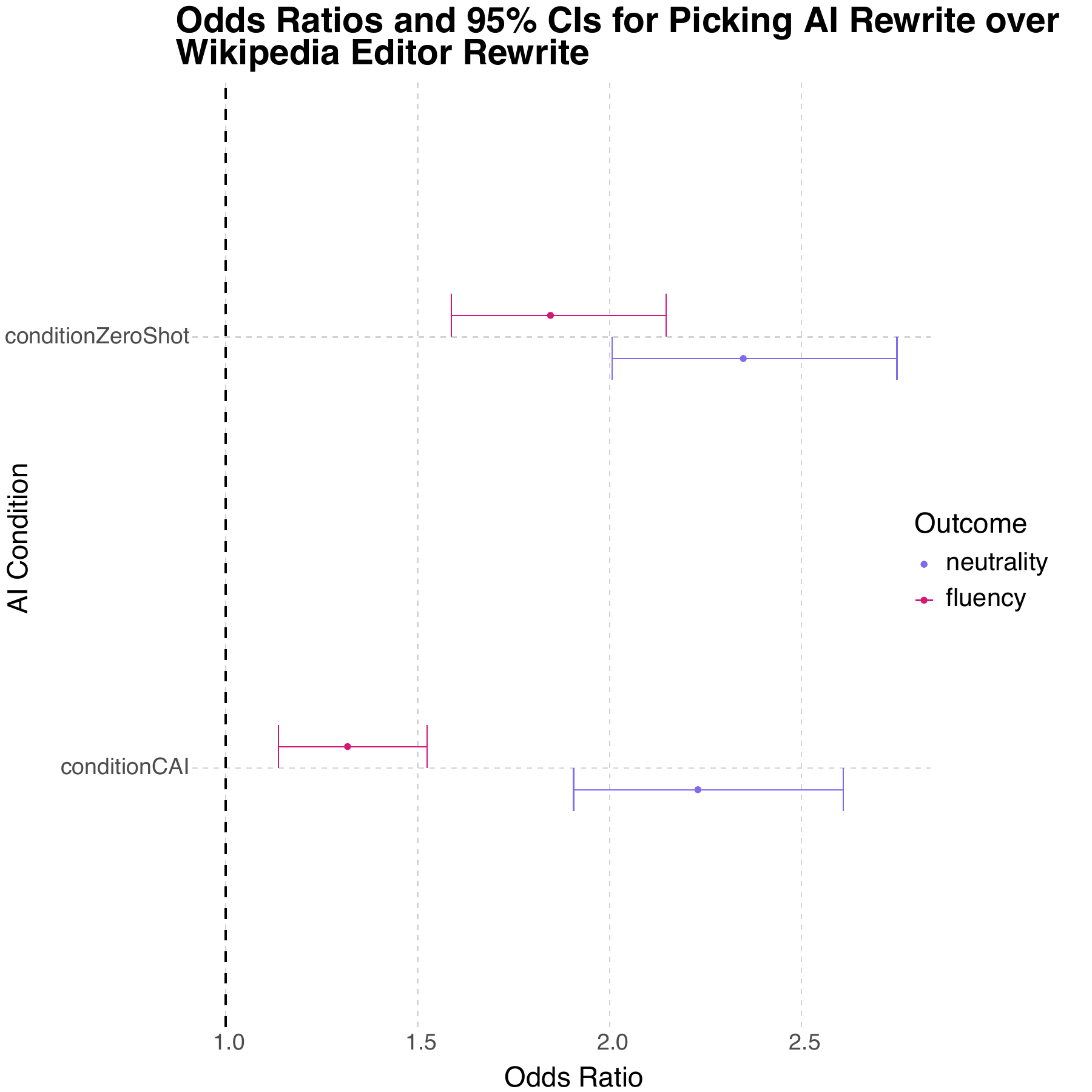}
        \caption{Participants selected AI edits as being more neutral and fluent than human edits. Odds ratios are from conditional logistic regression models (Appendix Figure \ref{con_neut_flu}).}
        \label{fig:ai_neutral_fluent}
    \end{subfigure}

    \begin{subfigure}[t]{0.48\textwidth}
        \centering
        \includegraphics[width=\linewidth]{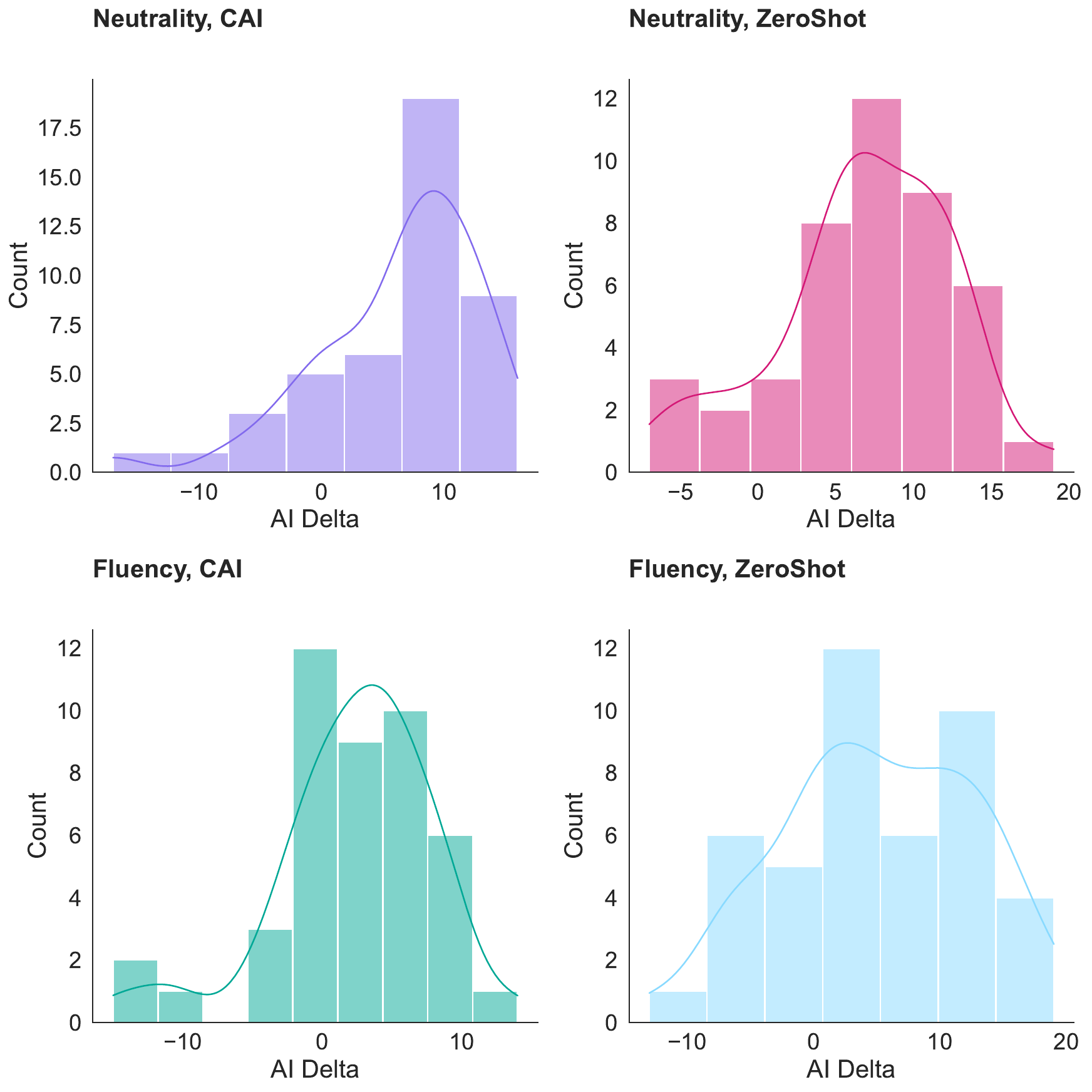}
        \caption{Distribution of AI Delta, where AI Delta is the matchup-level [number of raters who selected the AI edit] minus [the number of raters who selected the human edit].}
        \label{fig:ai_deltas}
    \end{subfigure}
    \caption{Experiment results for neutrality and fluency.}
    \label{fig:ai_human_comparison_expr}
\end{figure}

\paragraph{Analysis}
For answers involving choosing between two edits (fluency and neutrality), we first conducted binomial tests on whether the proportion of cases a participant chose the AI edit differed from 0.5. We also modeled participants' selections with a conditional logistic regression (\textit{clogit} function from the R package \textit{survival}), as is common for choice data~\cite{shang_dce_2023}. Analysis approaches agreed. We compared the frequencies of an edit adding or removing information via $\chi^2$ tests.

\subsection{Results}

\begin{table}
\footnotesize
\centering
\begin{tabular}{@{}l@{\hspace{4pt}}l@{\hspace{4pt}}l@{\hspace{4pt}}l@{\hspace{4pt}}l@{}}
\toprule
Question Type &        Comparison & AI & Human &        Delta \\
\midrule
   Neutrality & Human vs ZeroShot &    0.70 &       0.30 &     0.40**** \\
   Neutrality &      Human vs CAI &    0.69 &       0.31 &     0.38**** \\
      Fluency & Human vs ZeroShot &    0.65 &       0.35 &     0.30**** \\
      Fluency &      Human vs CAI &    0.57 &       0.43 &      0.14*** \\
          Add & Human vs ZeroShot &    0.27 &       0.28 & -0.02 (n.s.) \\
          Add &      Human vs CAI &    0.34 &       0.28 &       0.06** \\
       Remove & Human vs ZeroShot &    0.41 &       0.40 &  0.01 (n.s.) \\
       Remove &      Human vs CAI &    0.42 &       0.40 &  0.02 (n.s.) \\
\bottomrule
\end{tabular}
\caption{Experiment results. For fluency and neutrality, participants chose between an AI and a human edit. P-values are computed using two-tailed binomial tests for whether the probability of picking an AI edit differs from 0.5. For additions and removals, participants evaluated each of the AI and human edits separately but were shown both at the same time. We report human addition and removal data aggregated over both matchups. P-values are from chi-squared tests on whether human vs AI edits differed in frequencies of adding or removing information. `Delta' is the AI proportion minus the human proportion. Stars: n.s. $p>0.05, *p<0.05, **p<0.01 ***p<0.001, ****p<0.0001$}
\label{tab:sum_expr_stats_breakoutFalse}
\end{table}

\textit{\textbf{Label Quality.}} Participants (Appendix Table 11 for demographics) completed the task in a median of 19 minutes $(M = 21, SD = 9)$. The median correctness for the three training questions was 2 $(M=2, SD = 1)$. Agreement was high. The proportion of respondents agreeing with the majority answer for each question was: neutrality $(M = 0.74)$, fluency $(M = 0.69)$, additions $(M = 0.72)$, and removals $(M = 0.66)$. We conducted robustness checks (Appendix Table \ref{tab:sum_expr_stats_rob_check}) where we re-analyzed data after excluding participants with duration below the first quartile and including only those participants who got every training module question correct. Results were similar to the full dataset.

\textbf{\textit{Neutrality}.} See Table \ref{tab:sum_expr_stats_breakoutFalse} and Figure \ref{fig:ai_deltas} for raw data, Figure 4a for logistic regression odds ratios, and Appendix Table \ref{tab:sum_expr_stats_rob_check} for robustness checks. Participants rated AI edits as more neutral than human edits in 70\% of zero-shot choices (95\% CI = [67, 73]) and 69\% of CAI choices (95\% CI = [66, 72]), $p < 0.0001$ for both by two-tailed binomial tests and conditional logistic regressions. After our initial (surprising) results, we conducted a pilot study to rule out the possibility that the preference for AI edits was simply an artifact of forcing participants to choose between edits rather than allowing a ``both equal'' option. We find the proportional gap (probability of choosing the AI-generated edit over the human one) replicated, but in 21\% of cases, the options were rated as comparable (Appendix Table \ref{neut_pilot_nochoice}). The actual rate of AI neutrality preference (70\%) was generally higher than participants' forecasts $(M=58\%; SD = 16)$, but 75\% of participants predicted that AI edits would be chosen over 50\% of the time (Appendix Figure \ref{fig:neutrality_preds}).

\textbf{\textit{Fluency.}} Participants rated AI edits as more fluent (Table \ref{tab:sum_expr_stats_breakoutFalse}; Figure \ref{fig:ai_human_comparison_expr}) than human edits in 65\% of zero-shot choices (95\% CI = [61, 68]) and 57\% of CAI choices (95\% CI = [53, 60]), $p<0.001$ for both by two-tailed binomial tests and conditional logistic regression.

\textbf{\textit{Additions and removals.}} Participants evaluated whether each rewrite added or removed information from the original edit, excluding changes that increased NPOV compliance (Table \ref{tab:sum_expr_stats_breakoutFalse}). Participants saw two types of comparisons (or ``matchups''): human edits vs. zero-shot AI edits, and human edits vs. CAI edits. For each comparison, participants evaluated both edits separately. We analyzed judgments of human edits in two ways. When analyzing all human edit evaluations together (pooled across both matchup types), we found CAI edits added information more frequently (34\%, 95\% CI = [31, 38]) than human edits (28\%; 95\% CI = [26, 31]), $\chi^2$ $p<0.001$. However, when evaluating human edits alongside CAI edits, participants reported a higher information addition rate of the human edits (33\%, 95\% CI = [30, 37]) than when evaluating human edits alongside zero-shot edits (23\%; 95\% CI = [20, 26]). This led to a directionally similar but not statistically significant difference between human and CAI additions at the \textit{human vs CAI} matchup level (Appendix Table \ref{tab:sum_expr_stats_breakoutTrue}). High-addition CAI edits may have heightened participants' awareness or changed perceptions of additions in human edits.

\section{LLM Neutrality Generation: Qualitative Analysis}

To reconcile LLMs' low detection performance with crowdworkers' high preference for AI edits, we conducted a qualitative analysis of AI-generated vs human edits to better understand differences. We discuss cases (shown in Appendix Figure \ref{fig:cases}) that exemplify recurring patterns and significant implications that emerged upon inspection. 

AI is high-recall and low-precision compared to Wikipedia editors. Models generally change the text that editors change, but also much more (Cases 4, 5, and 9). For example, in one case (5) the human editor changed a single word (deleting the adjective ``unfortunately''). CAI and Zero-Shot made many changes (encompassing the human-editor one). LLMs make various grammatical and stylistic tweaks that may have influenced participants' judgments. We call the sum of these changes `NPOV+'.

In some cases, the AI model removed a string that a Wikipedia editor removed, but then replaced the string with one that was no better. In Case 1, Zero-Shot and the human editor essentially agreed on the problematic word (referencing a photographer as ``noted''). The CAI condition replaced ``noted'' with a long string that is just as NPOV-violating. 

In other cases (Cases 9 and 10), AI models arguably applied NPOV more faithfully than human editors, possibly due to varying community norms on what would constitute neutrality. For example, in Case 9, the AI models remove certain words like ``gimmick'' that human editors retained. In Case 10, the original text stated that an herbal medicine clinic ``can treat'' conditions. A Wikipedia editor changed this to ``treats'', suggesting greater certainty about herbal medicine's efficacy. The LLM edits expressed more skepticism: CAI modified it to ``claims to treat'' and Zero-Shot wrote ``provides services.'' Both stop short of asserting that the herbal medicine clinic \textit{definitively} treats conditions.

\section{Discussion}
LLMs are increasingly used in communities with their own rules and guidelines. Is providing these rules to LLMs enough for them to replicate community moderation decisions? We evaluated general-purpose LLMs on their ability to (1) detect edits that violated Wikipedia's Neutral Point of View (NPOV) and (2) generate NPOV-compliant edits from NPOV-violating edits. 

\paragraph{Takeaway 1: Large language models largely failed at detection.}
Our results provide insight into the capabilities and limitations of LLMs in applying community-specific neutrality norms. Across models and prompting strategies, performance was low---even with targeted prompts that directly incorporated Wikipedia's guidelines. LLMs are trained on Wikipedia data, so this poor performance existed despite a potential data leakage. Low detection performance suggests that applying subtle guidelines to real-world cases is difficult for today's general-purpose models. Notably, the largest model did the best. Perhaps even larger models would do better. Alternatively, low performance may be due to a more fundamental aspect of LLMs. For example, it may be that LLMs effectively `over-learn' a notion (e.g., neutrality) from broad web corpora, and specializing this notion to a community's norms requires changing a model's parameters. In any case, our study suggests two future directions. First, fine-tuning models on Wikipedia NPOV decisions may improve performance. However, fine-tuning may also risk overfitting~\cite{ma_adapting_2024}. Second, retrieval-augmented-generation (RAG) can incorporate community information (e.g., article Talk pages) to improve performance.

However, we must view LLM performance in the context of what is a hard task. NPOV edits are typically made by senior editors~\cite{pryzant_automatically_2020}. And crowdworkers could only guess the biased word in a Wikipedia edit 37\% of the time~\cite{recasens_linguistic_2013}. General-purpose LLMs might be better than random individuals at NPOV detection but worse than expert editors.

\paragraph{Takeaway 2: Different models had their own biases.}
Different models exhibited distinct biases. ChatGPT 3.5 over-predicted edits to be biased, while Mistral-Medium had the opposite tendency. Contrasting failure modes suggest that pre-trained LLMs may internalize distinct priors about what constitutes neutrality. These distinctions highlight one other shortcoming of relying on LLMs for such tasks. Models are introduced and updated rapidly, and our understanding of how LLMs work can get outdated just as rapidly. As language models become more ubiquitous, understanding these idiosyncratic biases will be crucial. Beyond understanding these biases, they may also be \textit{utilized}. Considering Wikipedia NPOV specifically, editors engaging with opposing views play a role in maintaining neutrality~\cite{greenstein_ideology_2021}. Perhaps multi-agent systems simulating pluralistic deliberation~\cite{10.1145/3706598.3713675} can better apply nuanced rules than individual agents. Our paper can serve as a roadmap for evaluating new LLMs and multi-agent systems on NPOV. 

\paragraph{Takeaway 3: LLMs may apply rules in different ways from humans.}
AI editors neutralize text differently from human editors. Computational metrics indicate that LLMs typically remove the NPOV-violating words that human editors remove (79\% average recall), but they also make many other changes (37\% average precision). Furthermore, while Wikipedians were more likely to neutralize text through deletions than additions, AI editors showed the opposite pattern. This divergence suggests LLMs may internalize and operationalize rules like NPOV in fundamentally different ways compared to a community's human experts. Though \textit{different} is not necessarily \textit{better} or \textit{worse}. AI moderators may reduce the cohesiveness of community content. But LLMs may also inspire new practices. More extensive editing from LLMs could be steered in beneficial directions or used when a ``severe'' NPOV violation is flagged. Future work can explore uses of LLMs for content moderation that go beyond attempting to `mimic' human content moderators. Alternatively, LLMs could generate \textit{multiple} neutralizations and then editors choose the most appealing.

\paragraph{Takeaway 4: Crowdworkers preferred LLM neutralizations to Wikipedia-editor neutralizations.}

Crowdworkers prefer AI edits over human edits on both fluency (61\% of choices) and neutrality (70\% of choices). We note that participants were \textit{not} Wikipedia editors. Their judgment may be more representative of Wikipedia readers. We hypothesize that our human evaluation findings are driven by both model size and reinforcement learning from human feedback (RLHF). First, LLMs have been trained on large corpora to predict likely tokens. This tendency to produce ``expected'' text (i.e., no grammar errors, standard language) may have caused participants to evaluate LLM generations more favorably. Relatedly, LLMs have been trained through RLHF to produce text that people---in general---will like~\cite{kaufmann_survey_2024}. And when evaluated by crowdworkers, their generations are preferred to Wikipedia-editor generations. Instruction-tuned models may apply rules in ways that resonate with a broader public, even if these applications differ from community experts. This suggests there may be potential trade-offs in satisfying Wikipedia readers and contributors (or more generally, the public and community experts). Fine-tuning models on expert feedback can increase alignment with domain specialists.

\paragraph{Takeaway 5: Qualitative analysis showed LLMs are `NPOV+'. }
To reconcile LLMs' low detection performance with high generation evaluations, we conducted a qualitative analysis of LLM generations. This analysis confirmed our computational findings: \textit{LLMs are high-recall, low-precision neutralizers}. We also found cases in which LLMs changed words human editors arguably \textit{should have changed.} If a community prioritizes recall over precision, then LLM generations are highly valuable.

However, LLMs also make many changes that are not always related to NPOV (e.g., readability). Participants rated CAI-generated rewrites as adding unnecessary information more frequently (34\%) than human rewrites (28\%). However, this difference was small compared to the large AI-human gap in changes measured by edit distance and diffs. Similarly, semantic analysis (Appendix~\ref{BertScore}) also showed that AI changes are relatively \textit{semantically} similar to Wikipedian changes. In triangulation, this suggests that LLMs typically make many unnecessary tweaks rather than change substantive information. A small internal annotation of LLM edits also qualitatively concurred: LLMs were not adding new facts, just unnecessary words (Appendix Table \ref{manual_annot_hal}). If deployed, volunteer editors may react negatively to AI systems making unnecessary edits to their content. This risks engendering a loss of agency, decreasing stylistic variation, and increasing moderator burden if moderators need to verify the correctness of changes. LLMs’ tendency to make extraneous edits necessitates human oversight to ensure LLMs do not overstep instructions.

\paragraph{Hybrid Human-AI Systems.}

Our findings suggest promise for LLMs in human-AI community moderation systems. LLMs performed better at generation than detection, so they can create `first drafts' of texts flagged by human moderators as norm-violating. However, humans flagging texts in the first place is labor-intensive. Future work could explore hybrid systems where models monitor Wikipedia Talk pages for comments suggesting NPOV violations. Based on automated triggers, these models could offer neutralizations. This hybrid approach would leverage indirect human input to address LLMs' detection shortfalls.  

\section{Limitations}
NPOV is a valuable test case since it is clearly articulated yet nuanced in application, and Wikipedia is widely read---though NPOV is just one community guideline. The WNC uses (a) single-editor judgments, (b) at one point in time (neutrality can change over time), (c) without full article context. We note that neutrality is an inherently subjective judgment and we rely on Wikipedia editors as a ground truth, though these editors may themselves disagree with each other---highlighting the challenge of real-world adjudication of community norms. Additionally, \citet{pryzant_automatically_2020} found 94.4\% of changes flagged as bias-related in WNC were in fact bias-related, also adding some measurement noise\footnote{We ran a Monte Carlo sensitivity analysis where for 1000 iterations, we randomly changed 5.6\% (estimate from \citet{pryzant_automatically_2020}) of biased labels to neutral, and then re-computed detection accuracy across all models and prompts. The average accuracy across these runs is 0.574 (SD = 0.002), similar to our observed overall accuracy of 0.581 (difference = 0.007), suggesting this relatively low amount of noise is not distorting conclusions.} to the dataset. Though if deployed in the wild, models would encounter the full range of edits flagged for NPOV concerns, including some noise. All of these factors make the dataset and task somewhat noisy. An ideal baseline would be a panel of Wikipedia editors (whose judgments are those that are used on Wikipedia), though such experts may be difficult to recruit. Our analysis of three prominent LLMs (GPT-4, ChatGPT 3.5, and Mistral-Medium) cannot generalize to \textit{all} LLMs. Similarly, we cannot rule out that some untested prompting strategy yields better detection results---though we explored various strategies including few-shot learning, chain-of-thought reasoning, self-optimizing prompts, official guidelines, and bias identification approaches. Future work can experiment with fine-tuning, multi-agent approaches, and RAG. Finally, our human evaluation relied on crowdworkers rather than Wikipedia editors. Crowdworkers are likely more representative of Wikipedia readers than editors, and edit preferences may differ between these groups. Hence, our human evaluation says more about lay readers than expert editor judgments. Despite these limitations, our work offers a detailed analysis of the NPOV detection and generation ability of general-purpose LLMs.

\section{Conclusion}

Exposing models to high-level principles alone was insufficient for replicating community members' judgments in specific cases. When models correctly classified edits, this was often linked to a single ``giveaway'' adjective. However, LLMs successfully applied NPOV in generation, mirroring changes made by Wikipedia editors and being preferred by laypeople. This suggests a gap between following general instructions and applying them to specific cases as community members do. Even when articulating principles is easy, applying them to particular cases---like community members do---remains challenging for LLMs.

\section*{Ethics Statement}

There are ethical considerations to deploying LLMs for NPOV. Readers trust Wikipedia partially due to Wikipedia's editorial process~\cite{elmimouni_why_2022}. Some evidence suggests users trust ChatGPT less than Wikipedia~\cite{jung_we_2024}, so LLM-generated content risks reducing trust in Wikipedia. Also, initial negative feedback demotivates newcomers in peer production~\cite{halfaker_dont_2011}---hence, extensive LLM rewrites may harm community development. 

\bibliography{references_clean}

\begin{thebibliography}{43}
\providecommand{\natexlab}[1]{#1}

\bibitem[{Ashkinaze et~al.(2025)Ashkinaze, Fry, Edara, Gilbert, and Budak}]{10.1145/3706598.3713675}
Ashkinaze, J.; Fry, E.; Edara, N.; Gilbert, E.; and Budak, C. 2025.
\newblock Plurals: A System for Guiding LLMs via Simulated Social Ensembles.
\newblock In \emph{Proceedings of the 2025 CHI Conference on Human Factors in Computing Systems}, CHI '25. New York, NY, USA: Association for Computing Machinery.
\newblock ISBN 9798400713941.

\bibitem[{Bai et~al.(2022)Bai, Kadavath, Kundu, Askell, Kernion, Jones, Chen, Goldie, Mirhoseini, McKinnon, Chen, Olsson, Olah, Hernandez, Drain, Ganguli, Li, Tran-Johnson, Perez, Kerr, Mueller, Ladish, Landau, Ndousse, Lukosuite, Lovitt, Sellitto, Elhage, Schiefer, Mercado, DasSarma, Lasenby, Larson, Ringer, Johnston, Kravec, Showk, Fort, Lanham, Telleen-Lawton, Conerly, Henighan, Hume, Bowman, Hatfield-Dodds, Mann, Amodei, Joseph, McCandlish, Brown, and Kaplan}]{bai_constitutional_2022}
Bai, Y.; Kadavath, S.; Kundu, S.; Askell, A.; Kernion, J.; Jones, A.; Chen, A.; Goldie, A.; Mirhoseini, A.; McKinnon, C.; Chen, C.; Olsson, C.; Olah, C.; Hernandez, D.; Drain, D.; Ganguli, D.; Li, D.; Tran-Johnson, E.; Perez, E.; Kerr, J.; Mueller, J.; Ladish, J.; Landau, J.; Ndousse, K.; Lukosuite, K.; Lovitt, L.; Sellitto, M.; Elhage, N.; Schiefer, N.; Mercado, N.; DasSarma, N.; Lasenby, R.; Larson, R.; Ringer, S.; Johnston, S.; Kravec, S.; Showk, S.~E.; Fort, S.; Lanham, T.; Telleen-Lawton, T.; Conerly, T.; Henighan, T.; Hume, T.; Bowman, S.~R.; Hatfield-Dodds, Z.; Mann, B.; Amodei, D.; Joseph, N.; McCandlish, S.; Brown, T.; and Kaplan, J. 2022.
\newblock Constitutional {AI}: {Harmlessness} from {AI} {Feedback}.

\bibitem[{Butler, Joyce, and Pike(2008)}]{butler_dont_2008}
Butler, B.; Joyce, E.; and Pike, J. 2008.
\newblock Don't look now, but we've created a bureaucracy: the nature and roles of policies and rules in wikipedia.
\newblock In \emph{Proc. of the {SIGCHI} {Conf.} on {Human} {Factors} in {Computing} {Systems}}, {CHI} '08, 1101--1110. New York, NY, USA: Assoc. for Comput. Mach.
\newblock ISBN 978-1-60558-011-1.

\bibitem[{Cao et~al.(2024)Cao, Domingo, Gilbert, Mazurek, Shilton, and Daumé~III}]{cao_toxicity_2024}
Cao, Y.~T.; Domingo, L.-F.; Gilbert, S.~A.; Mazurek, M.; Shilton, K.; and Daumé~III, H. 2024.
\newblock Toxicity {Detection} is {NOT} all you {Need}: {Measuring} the {Gaps} to {Supporting} {Volunteer} {Content} {Moderators}.
\newblock arXiv:2311.07879.

\bibitem[{Elmimouni, Forte, and Morgan(2022)}]{elmimouni_why_2022}
Elmimouni, H.; Forte, A.; and Morgan, J. 2022.
\newblock Why {People} {Trust} {Wikipedia} {Articles}: {Credibility} {Assessment} {Strategies} {Used} by {Readers}.
\newblock In \emph{Proc. of the 18th {Int.} {Symp.} on {Open} {Collaboration}}, {OpenSym} '22, 1--10. New York, NY, USA: Assoc. for Comput. Mach.
\newblock ISBN 978-1-4503-9845-9.

\bibitem[{{FORCE11}(2020)}]{fair}
{FORCE11}. 2020.
\newblock The FAIR Data principles.
\newblock \url{https://force11.org/info/the-fair-data-principles/}.

\bibitem[{Gebru et~al.(2021)Gebru, Morgenstern, Vecchione, Vaughan, Wallach, Iii, and Crawford}]{gebru2021datasheets}
Gebru, T.; Morgenstern, J.; Vecchione, B.; Vaughan, J.~W.; Wallach, H.; Iii, H.~D.; and Crawford, K. 2021.
\newblock Datasheets for datasets.
\newblock \emph{Communications of the ACM}, 64(12): 86--92.

\bibitem[{Gilardi, Alizadeh, and Kubli(2023)}]{gilardi_chatgpt_2023}
Gilardi, F.; Alizadeh, M.; and Kubli, M. 2023.
\newblock {ChatGPT} outperforms crowd workers for text-annotation tasks.
\newblock \emph{Proc. of the National Academy of Sciences}, 120(30): e2305016120.

\bibitem[{Greenstein, Gu, and Zhu(2021)}]{greenstein_ideology_2021}
Greenstein, S.; Gu, G.; and Zhu, F. 2021.
\newblock Ideology and {Composition} {Among} an {Online} {Crowd}: {Evidence} from {Wikipedians}.
\newblock \emph{Management Science}, 67(5): 3067--3086.

\bibitem[{Halfaker and Geiger(2020)}]{halfaker_ores_2020}
Halfaker, A.; and Geiger, R.~S. 2020.
\newblock Ores: {Lowering} barriers with participatory machine learning in wikipedia.
\newblock \emph{Proc. of the ACM on Human-Computer Interaction}, 4(CSCW2): 1--37.
\newblock Publisher: ACM New York, NY, USA.

\bibitem[{Halfaker, Kittur, and Riedl(2011)}]{halfaker_dont_2011}
Halfaker, A.; Kittur, A.; and Riedl, J. 2011.
\newblock Don't bite the newbies: how reverts affect the quantity and quality of {Wikipedia} work.
\newblock In \emph{Proc. of the 7th {Int.} {Symp.} on {Wikis} and {Open} {Collaboration}}, {WikiSym} '11, 163--172. New York, NY, USA: Assoc. for Comput. Mach.
\newblock ISBN 978-1-4503-0909-7.

\bibitem[{Hansen, Berente, and Lyytinen(2009)}]{hansen_wikipedia_2009}
Hansen, S.; Berente, N.; and Lyytinen, K. 2009.
\newblock Wikipedia, {Critical} {Social} {Theory}, and the {Possibility} of {Rational} {Discourse} 1.
\newblock \emph{The Information Society}, 25(1): 38--59.

\bibitem[{Harrison(2023)}]{harrison_should_2023}
Harrison, S. 2023.
\newblock Should {ChatGPT} {Be} {Used} to {Write} {Wikipedia} {Articles}?
\newblock \emph{Slate}.

\bibitem[{Jung et~al.(2024)Jung, Chen, Jang, and Sundar}]{jung_we_2024}
Jung, Y.; Chen, C.; Jang, E.; and Sundar, S.~S. 2024.
\newblock Do {We} {Trust} {ChatGPT} as much as {Google} {Search} and {Wikipedia}?
\newblock In \emph{Extended {Abstracts} of the {CHI} {Conf.} on {Human} {Factors} in {Computing} {Systems}}, {CHI} {EA} '24, 1--9. New York, NY, USA: Assoc. for Comput. Mach.
\newblock ISBN 9798400703317.

\bibitem[{Kaufmann et~al.(2024)Kaufmann, Weng, Bengs, and Hüllermeier}]{kaufmann_survey_2024}
Kaufmann, T.; Weng, P.; Bengs, V.; and Hüllermeier, E. 2024.
\newblock A {Survey} of {Reinforcement} {Learning} from {Human} {Feedback}.
\newblock arXiv:2312.14925.

\bibitem[{Khattab et~al.(2023)Khattab, Singhvi, Maheshwari, Zhang, Santhanam, Vardhamanan, Haq, Sharma, Joshi, Moazam, Miller, Zaharia, and Potts}]{khattab_dspy_2023}
Khattab, O.; Singhvi, A.; Maheshwari, P.; Zhang, Z.; Santhanam, K.; Vardhamanan, S.; Haq, S.; Sharma, A.; Joshi, T.~T.; Moazam, H.; Miller, H.; Zaharia, M.; and Potts, C. 2023.
\newblock {DSPy}: {Compiling} {Declarative} {Language} {Model} {Calls} into {Self}-{Improving} {Pipelines}.

\bibitem[{Kittur and Kraut(2008)}]{kittur_harnessing_2008}
Kittur, A.; and Kraut, R.~E. 2008.
\newblock Harnessing the wisdom of crowds in wikipedia: quality through coordination.
\newblock In \emph{Proc. of the 2008 {ACM} Conf. on {Computer} supported cooperative work}, {CSCW} '08, 37--46. New York, NY, USA: Assoc. for Comput. Mach.
\newblock ISBN 978-1-60558-007-4.

\bibitem[{Kittur et~al.(2007)Kittur, Suh, Pendleton, and Chi}]{kittur_he_2007}
Kittur, A.; Suh, B.; Pendleton, B.~A.; and Chi, E.~H. 2007.
\newblock He says, she says: conflict and coordination in {Wikipedia}.
\newblock In \emph{Proc. of the {SIGCHI} {Conf.} on {Human} {Factors} in {Computing} {Systems}}, 453--462. San Jose California USA: ACM.
\newblock ISBN 978-1-59593-593-9.

\bibitem[{Kolla et~al.(2024)Kolla, Salunkhe, Chandrasekharan, and Saha}]{kolla_llm-mod_2024}
Kolla, M.; Salunkhe, S.; Chandrasekharan, E.; and Saha, K. 2024.
\newblock {LLM}-{Mod}: {Can} {Large} {Language} {Models} {Assist} {Content} {Moderation}?
\newblock In \emph{Extended {Abstracts} of the 2024 {CHI} {Conf.} on {Human} {Factors} in {Computing} {Systems}}, {CHI} {EA} '24, 1--8. New York, NY, USA: Assoc. for Comput. Mach.
\newblock ISBN 9798400703317.

\bibitem[{Kumar, AbuHashem, and Durumeric(2024)}]{kumar_watch_2024}
Kumar, D.; AbuHashem, Y.~A.; and Durumeric, Z. 2024.
\newblock Watch {Your} {Language}: {Investigating} {Content} {Moderation} with {Large} {Language} {Models}.
\newblock \emph{Proc. of the Int. AAAI Conf. on Web and Social Media}, 18: 865--878.

\bibitem[{Lin and Och(2004)}]{lin_orange_2004}
Lin, C.-Y.; and Och, F.~J. 2004.
\newblock {ORANGE}: a {Method} for {Evaluating} {Automatic} {Evaluation} {Metrics} for {Machine} {Translation}.
\newblock In \emph{{COLING} 2004: {Proceedings} of the 20th {Int.} {Conf.} on {Computational} {Linguistics}}, 501--507. Geneva, Switzerland: COLING.

\bibitem[{Ma et~al.(2024)Ma, Zhang, Fu, Zhao, and Wu}]{ma_adapting_2024}
Ma, H.; Zhang, C.; Fu, H.; Zhao, P.; and Wu, B. 2024.
\newblock Adapting {Large} {Language} {Models} for {Content} {Moderation}: {Pitfalls} in {Data} {Engineering} and {Supervised} {Fine}-tuning.

\bibitem[{Madanagopal and Caverlee(2022)}]{madanagopal_improving_2022}
Madanagopal, K.; and Caverlee, J. 2022.
\newblock Improving linguistic bias detection in Wikipedia using cross-domain adaptive pre-training.
\newblock In \emph{Companion Proceedings of the Web Conference 2022}, 1301--1309.

\bibitem[{Madanagopal and Caverlee(2023{\natexlab{a}})}]{madanagopal_bias_2023}
Madanagopal, K.; and Caverlee, J. 2023{\natexlab{a}}.
\newblock Bias neutralization in non-parallel texts: A cyclic approach with auxiliary guidance.
\newblock In \emph{Proceedings of the 2023 Conference on Empirical Methods in Natural Language Processing}, 14265--14278.

\bibitem[{Madanagopal and Caverlee(2023{\natexlab{b}})}]{madanagopal_reinforced_2023}
Madanagopal, K.; and Caverlee, J. 2023{\natexlab{b}}.
\newblock Reinforced sequence training based subjective bias correction.
\newblock In \emph{Proceedings of the 17th Conference of the European Chapter of the Association for Computational Linguistics}, 2585--2598.

\bibitem[{McDowell and Vetter(2020)}]{mcdowell_it_2020}
McDowell, Z.~J.; and Vetter, M.~A. 2020.
\newblock It {Takes} a {Village} to {Combat} a {Fake} {News} {Army}: {Wikipedia}’s {Community} and {Policies} for {Information} {Literacy}.
\newblock \emph{Social Media + Society}, 6(3): 2056305120937309.

\bibitem[{Orme(1998)}]{orme_sample_1998}
Orme, B. 1998.
\newblock Sample size issues for conjoint analysis studies.
\newblock \emph{Sequim: Sawtooth Software Technical Paper}.

\bibitem[{Papineni et~al.(2002)Papineni, Roukos, Ward, and Zhu}]{papineni_bleu_2002}
Papineni, K.; Roukos, S.; Ward, T.; and Zhu, W.-J. 2002.
\newblock Bleu: a {Method} for {Automatic} {Evaluation} of {Machine} {Translation}.
\newblock In Isabelle, P.; Charniak, E.; and Lin, D., eds., \emph{Proc. of the 40th {Annual} {Meeting} of the {Association} for {Computational} {Linguistics}}, 311--318. Philadelphia, Pennsylvania, USA: Assoc. for Computational Linguistics.

\bibitem[{Pryzant et~al.(2020)Pryzant, Diehl~Martinez, Dass, Kurohashi, Jurafsky, and Yang}]{pryzant_automatically_2020}
Pryzant, R.; Diehl~Martinez, R.; Dass, N.; Kurohashi, S.; Jurafsky, D.; and Yang, D. 2020.
\newblock Automatically {Neutralizing} {Subjective} {Bias} in {Text}.
\newblock \emph{Proc. of the AAAI Conf. on Artificial Intelligence}, 34(01): 480--489.

\bibitem[{Rawat et~al.(2019)Rawat, Sarkar, Singh, Alvarado, and Rasberry}]{rawat_automatic_2019}
Rawat, C.; Sarkar, A.; Singh, S.; Alvarado, R.; and Rasberry, L. 2019.
\newblock Automatic detection of online abuse and analysis of problematic users in wikipedia.
\newblock In \emph{2019 {Systems} and {Information} {Engineering} {Design} {Symp.} ({SIEDS})}, 1--6. IEEE.

\bibitem[{Reagle(2007)}]{reagle_is_2007}
Reagle, J. 2007.
\newblock Is the {Wikipedia} {Neutral}?

\bibitem[{Recasens, Danescu-Niculescu-Mizil, and Jurafsky(2013)}]{recasens_linguistic_2013}
Recasens, M.; Danescu-Niculescu-Mizil, C.; and Jurafsky, D. 2013.
\newblock Linguistic models for analyzing and detecting biased language.
\newblock In \emph{Proc. of the 51st annual meeting of the {Association} for {Computational} {Linguistics} (volume 1: long papers)}, 1650--1659.

\bibitem[{Scott(1998)}]{scott_seeing_1998}
Scott, J.~C. 1998.
\newblock \emph{Seeing like a state : how certain schemes to improve the human condition have failed}.
\newblock Yale agrarian studies. Yale University Press.

\bibitem[{Shang and Chandra(2023)}]{shang_dce_2023}
Shang, L.; and Chandra, Y. 2023.
\newblock {DCE} {Data} {Analysis} {Using} {R}.
\newblock In Shang, L.; and Chandra, Y., eds., \emph{Discrete {Choice} {Experiments} {Using} {R}: {A} {How}-{To} {Guide} for {Social} and {Managerial} {Sciences}}, 157--181. Singapore: Springer Nature.
\newblock ISBN 978-981-9945-62-7.

\bibitem[{Steinsson(2024)}]{steinsson_rule_2024}
Steinsson, S. 2024.
\newblock Rule {Ambiguity}, {Institutional} {Clashes}, and {Population} {Loss}: {How} {Wikipedia} {Became} the {Last} {Good} {Place} on the {Internet}.
\newblock \emph{American Political Science Review}, 118(1): 235--251.

\bibitem[{Suchman(1987)}]{suchman_plans_1987}
Suchman, L.~A. 1987.
\newblock \emph{Plans and situated actions: the problem of human-machine communication}.
\newblock USA: Cambridge University Press.
\newblock ISBN 978-0-521-33137-1.

\bibitem[{Swarts(2009)}]{swarts_collaborative_2009}
Swarts, J. 2009.
\newblock The collaborative construction of "fact" on {Wikipedia}.
\newblock In \emph{Proc. of the 27th {ACM} Int. Conf. on {Design} of communication}, 281--288. Bloomington Indiana USA: ACM.
\newblock ISBN 978-1-60558-559-8.

\bibitem[{Wales(2006)}]{wales_jimmy_2006}
Wales, J. 2006.
\newblock Jimmy {Wales}: {The} birth of {Wikipedia} {\textbar} {TED} {Talk}.

\bibitem[{Wang et~al.(2020)Wang, Yao, Kwok, and Ni}]{wang_generalizing_2020}
Wang, Y.; Yao, Q.; Kwok, J.~T.; and Ni, L.~M. 2020.
\newblock Generalizing from a {Few} {Examples}: {A} {Survey} on {Few}-shot {Learning}.
\newblock \emph{ACM Comput. Surv.}, 53(3): 63:1--63:34.

\bibitem[{Wei et~al.(2023)Wei, Wang, Schuurmans, Bosma, Ichter, Xia, Chi, Le, and Zhou}]{wei_chain--thought_2023}
Wei, J.; Wang, X.; Schuurmans, D.; Bosma, M.; Ichter, B.; Xia, F.; Chi, E.; Le, Q.; and Zhou, D. 2023.
\newblock Chain-of-{Thought} {Prompting} {Elicits} {Reasoning} in {Large} {Language} {Models}.
\newblock arXiv:2201.11903.

\bibitem[{Wikipedia(2023)}]{wikipedia_core_2023}
Wikipedia. 2023.
\newblock Core content policies.
\newblock \url{https://w.wiki/CjbT}.

\bibitem[{Zhang et~al.(2020)Zhang, Kishore, Wu, Weinberger, and Artzi}]{zhang2020bertscoreevaluatingtextgeneration}
Zhang, T.; Kishore, V.; Wu, F.; Weinberger, K.~Q.; and Artzi, Y. 2020.
\newblock BERTScore: Evaluating Text Generation with BERT.
\newblock arXiv:1904.09675.

\bibitem[{Zhong et~al.(2021)Zhong, Yang, Xu, and Yang}]{zhong_wikibias_2021}
Zhong, Y.; Yang, J.; Xu, W.; and Yang, D. 2021.
\newblock {WIKIBIAS}: {Detecting} {Multi}-{Span} {Subjective} {Biases} in {Language}.
\newblock In Moens, M.-F.; Huang, X.; Specia, L.; and Yih, S. W.-t., eds., \emph{Findings of the {Association} for {Computational} {Linguistics}: {EMNLP} 2021}, 1799--1814. Punta Cana, Dominican Republic: Association for Computational Linguistics.

\end{thebibliography}

\section*{Paper Checklist}

\newcommand{\answerYes}[1]{\textcolor{blue}{#1}} 
\newcommand{\answerNo}[1]{\textcolor{teal}{#1}} 
\newcommand{\answerNA}[1]{\textcolor{gray}{#1}} 
\newcommand{\answerTODO}[1]{\textcolor{red}{#1}} 

\lstset{%
	basicstyle={\footnotesize\ttfamily},
	numbers=left,numberstyle=\footnotesize,xleftmargin=2em,
	aboveskip=0pt,belowskip=0pt,%
	showstringspaces=false,tabsize=2,breaklines=true}
\floatstyle{ruled}
\newfloat{listing}{tb}{lst}{}
\floatname{listing}{Listing}

\begin{enumerate}

\item For most authors...
\begin{enumerate}
    \item  Would answering this research question advance science without violating social contracts, such as violating privacy norms, perpetuating unfair profiling, exacerbating the socio-economic divide, or implying disrespect to societies or cultures?
    \answerYes{Yes, we focus on enhancing the neutrality of online content.}
  \item Do your main claims in the abstract and introduction accurately reflect the paper's contributions and scope?
    \answerYes{Yes, See Discussion section.}
   \item Do you clarify how the proposed methodological approach is appropriate for the claims made? 
    \answerYes{Yes, each experiment has an Experiment Setup section.}
   \item Do you clarify what are possible artifacts in the data used, given population-specific distributions?
    \answerYes{Yes, this is discussed in the Limitations.}
  \item Did you describe the limitations of your work?
    \answerYes{Yes, this is discussed in the Limitations.}
  \item Did you discuss any potential negative societal impacts of your work?
    \answerYes{Yes, See Ethics Statement.}
      \item Did you discuss any potential misuse of your work?
   \answerYes{Yes, See Ethics Statement.}
    \item Did you describe steps taken to prevent or mitigate potential negative outcomes of the research, such as data and model documentation, data anonymization, responsible release, access control, and the reproducibility of findings?
    \answerNA{NA.}
  \item Have you read the ethics review guidelines and ensured that your paper conforms to them?
    \answerYes{Yes.}
\end{enumerate}

\item Additionally, if your study involves hypotheses testing...
\begin{enumerate}
  \item Did you clearly state the assumptions underlying all theoretical results?
    \answerNA{NA}
  \item Have you provided justifications for all theoretical results?
   \answerNA{NA}
  \item Did you discuss competing hypotheses or theories that might challenge or complement your theoretical results?
   \answerNA{NA}
  \item Have you considered alternative mechanisms or explanations that might account for the same outcomes observed in your study?
   \answerNA{NA}
  \item Did you address potential biases or limitations in your theoretical framework?
   \answerNA{NA}
  \item Have you related your theoretical results to the existing literature in social science?
   \answerNA{NA}
  \item Did you discuss the implications of your theoretical results for policy, practice, or further research in the social science domain?
    \answerNA{NA}
\end{enumerate}

\item Additionally, if you are including theoretical proofs...
\begin{enumerate}
  \item Did you state the full set of assumptions of all theoretical results?
    \answerNA{NA}
	\item Did you include complete proofs of all theoretical results?
    \answerNA{NA}
\end{enumerate}

\item Additionally, if you ran machine learning experiments...
\begin{enumerate}
  \item Did you include the code, data, and instructions needed to reproduce the main experimental results (either in the supplemental material or as a URL)?
    \answerNA{NA}
  \item Did you specify all the training details (e.g., data splits, hyperparameters, how they were chosen)?
    \answerNA{NA}
     \item Did you report error bars (e.g., with respect to the random seed after running experiments multiple times)?
    \answerNA{NA}
	\item Did you include the total amount of compute and the type of resources used (e.g., type of GPUs, internal cluster, or cloud provider)?
    \answerNA{NA}
     \item Do you justify how the proposed evaluation is sufficient and appropriate to the claims made? 
    \answerNA{NA}
     \item Do you discuss what is ``the cost`` of misclassification and fault (in)tolerance?
    \answerNA{NA}
  
\end{enumerate}

\item Additionally, if you are using existing assets (e.g., code, data, models) or curating/releasing new assets, \textbf{without compromising anonymity}...
\begin{enumerate}
  \item If your work uses existing assets, did you cite the creators?
    \answerYes{Yes, see Dataset section.}
  \item Did you mention the license of the assets?
    \answerYes{Yes, see Dataset section.}
  \item Did you include any new assets in the supplemental material or as a URL?
    \answerNo{No.}
  \item Did you discuss whether and how consent was obtained from people whose data you're using/curating?
    \answerNo{No, the dataset we used consisted of publicly available Wikipedia edits.}
  \item Did you discuss whether the data you are using/curating contains personally identifiable information or offensive content?
    \answerNA{NA, the WNC is indexed by IDs of edits and not IDs of editors---though the latter (which is publically available) can be obtained from the former.}
\item If you are curating or releasing new datasets, did you discuss how you intend to make your datasets FAIR (see \citet{fair})?
\answerNA{NA}
\item If you are curating or releasing new datasets, did you create a Datasheet for the Dataset (see \citet{gebru2021datasheets})? 
\answerNA{NA}
\end{enumerate}

\item Additionally, if you used crowdsourcing or conducted research with human subjects, \textbf{without compromising anonymity}...
\begin{enumerate}
  \item Did you include the full text of instructions given to participants and screenshots?
    \answerYes{Yes, see Supplementary Materials file.}
  \item Did you describe any potential participant risks, with mentions of Institutional Review Board (IRB) approvals?
    \answerYes{Yes.}
  \item Did you include the estimated hourly wage paid to participants and the total amount spent on participant compensation?
    \answerYes{Yes, see LLM Neutrality Generation: Human Evaluation section.}
   \item Did you discuss how data is stored, shared, and deidentified?
   \answerYes{Data will be stored on secure servers.}
\end{enumerate}

\end{enumerate}

\appendix

\begin{figure*}[t!]
  \section{Qualitative Analysis of Edits}
  \centering
  \begin{subfigure}[t]{0.4\textwidth}
      \centering
      \caption{}
      \includegraphics[width=\textwidth]{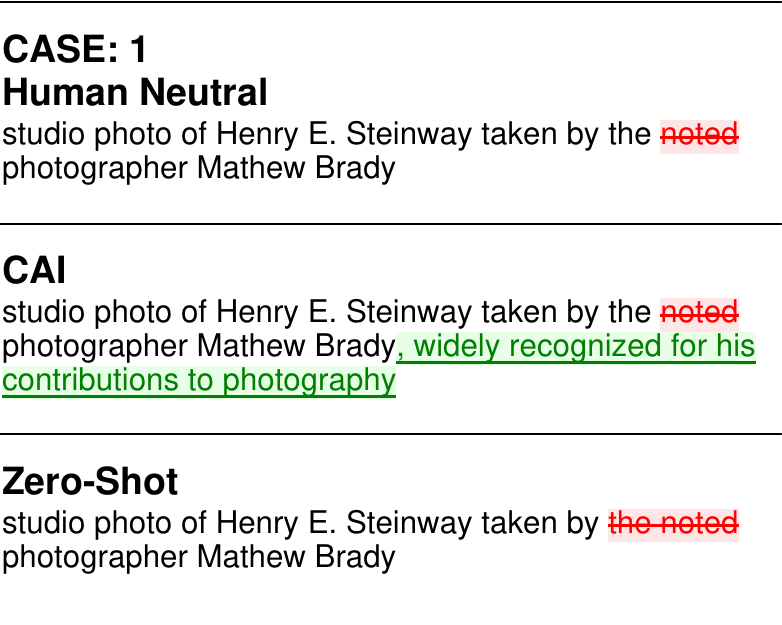}
      \label{fig:case1}
  \end{subfigure}
  \hfill
  \begin{subfigure}[t]{0.4\textwidth}
      \centering
      \caption{}
      \includegraphics[width=\textwidth]{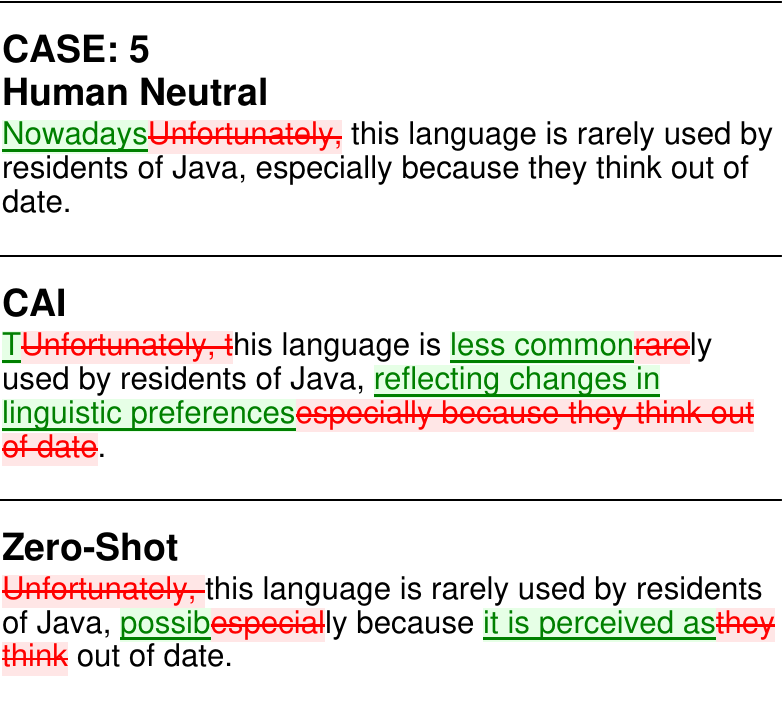}
      \label{fig:case5}
  \end{subfigure}
  \vspace{-0.5cm}
  \begin{subfigure}[t]{0.4\textwidth}
      \centering
      \caption{}
      \includegraphics[width=\textwidth]{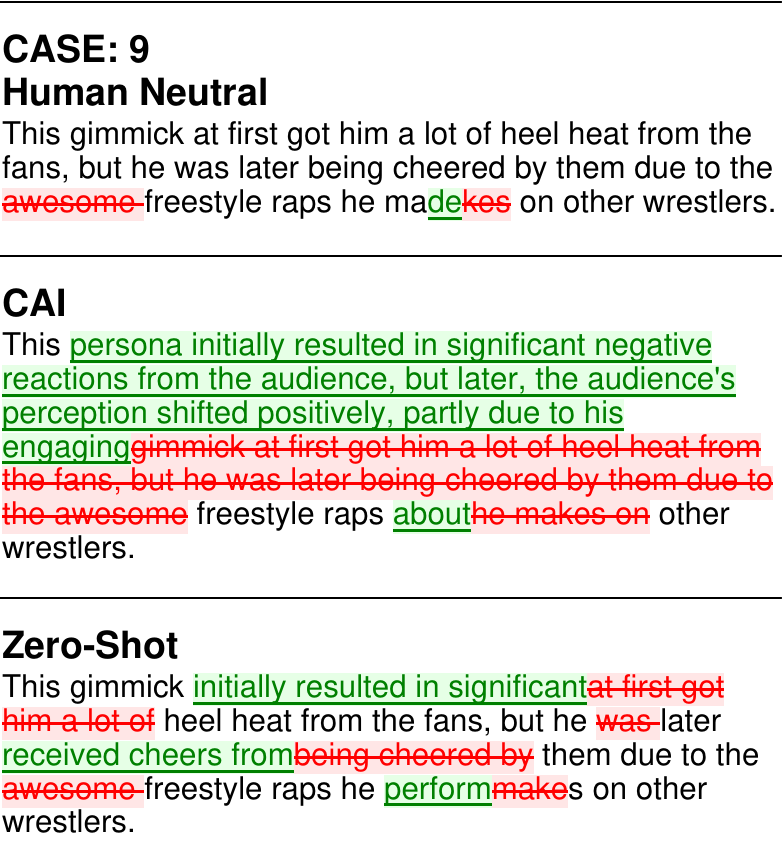}
      \label{fig:case9}
  \end{subfigure}
  \hfill
  \begin{subfigure}[t]{0.4\textwidth}
      \centering
      \caption{}
      \includegraphics[width=\textwidth]{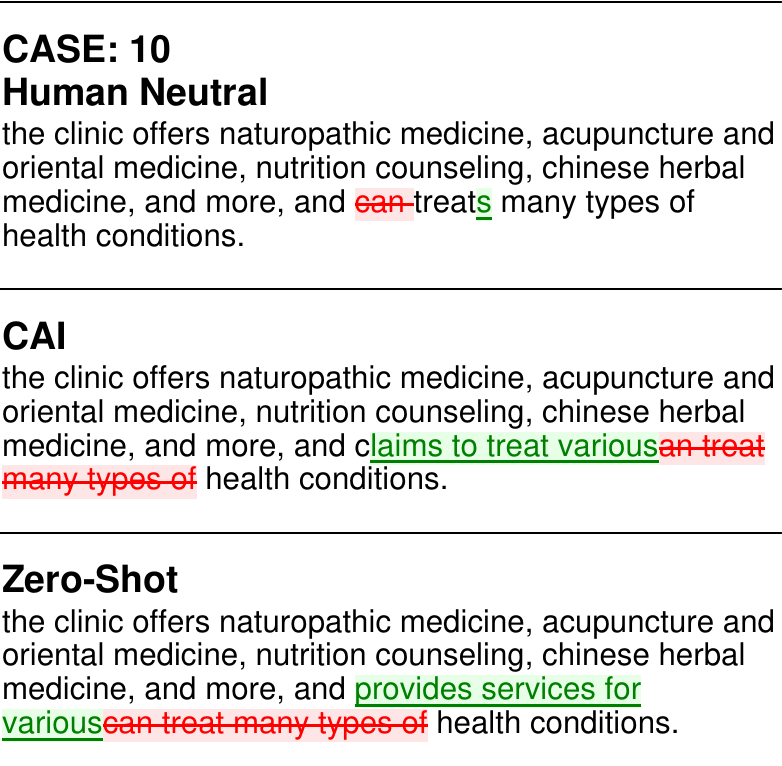}
      \label{fig:case10}
  \end{subfigure}
  \vspace{-0.5cm}
  \begin{subfigure}[t]{0.85\textwidth}
      \centering
      \caption{}
      \includegraphics[width=\textwidth]{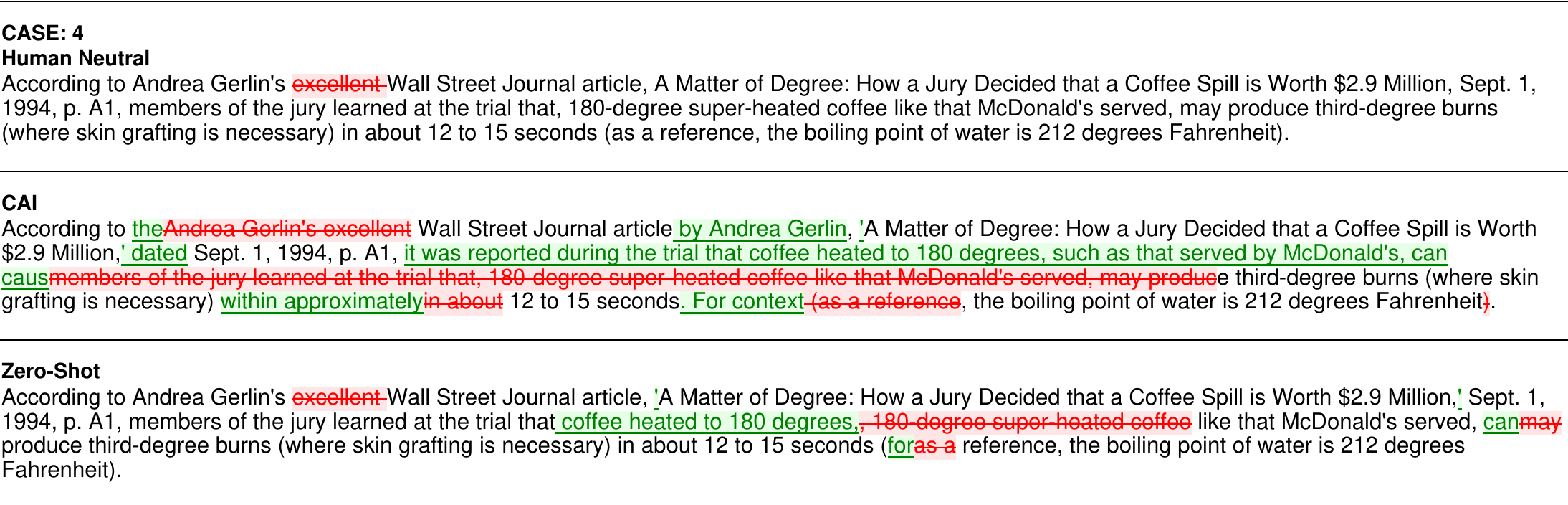}
      \label{fig:case4}
  \end{subfigure}
  \caption{Comparing AI neutralizations to Wikipedia editor neutralizations.}
  \label{fig:cases}
\end{figure*}
\clearpage

\subsection{Internal Annotation of Edits}

\begin{table}[h!]
\begin{tabular}{lll}
\toprule
Topic & Additions & Removals \\
\midrule
Culture & 3/12 & 3/12 \\
Geography & 6/12 & 0/12 \\
History And Society & 1/12 & 0/12 \\
STEM & 4/12 & 0/12 \\
\bottomrule
\end{tabular}
\caption{Two members of the study team manually annotated 48 edits for additions and removals, paying particular to attention to whether LLMs hallucinated.}
\label{manual_annot_hal}
\end{table}

While our human experiment is higher powered and represents more diverse perspectives, two members of the study team manually annotated 48 edits (12 per topic) using consensus coding for additions and removals (Table \ref{manual_annot_hal}), using the same task instructions as were given to study participants. Qualitatively, our observation was that the most common form of unnecessary additions was unnecessary both-sideism or hedges. As an example, one edit discussed the availability of a feature across operating systems, and the AI editor (needlessly) added ``It is important to note that the level of support can vary and should be verified with specific operating system documentation or sources.'' We did not find any hallucination of \textit{facts} in the 48 edits we annotated. The closest was that for an edit about evolution, an AI editor added that evolution ``is supported by a vast body of evidence and widely accepted within the scientific community.''---which is true, but was not necessary for neutralizing the edit.

\section{Prompts}
\label{appendix_prompts}

\subsection{Detection: NPOV}
\textit{Note: The EXAMPLES part of the prompt is only included for the few-shot conditions.}

\begin{scriptsize}
\begin{verbatim}
NSTRUCTIONS
Act like an expert Wikipedia editor. Classify if
this Wikipedia edit is neutral or biased.

A Wikipedia edit is biased if it violates
Wikipedia’s neutral
point of view (NPOV) policy.

Neutral point of view policy:
- Policy 1: Avoid stating opinions as facts.
- Policy 2: Avoid stating seriously contested
assertions as facts.
- Policy 3: Avoid stating facts as opinions.
- Policy 4: Prefer nonjudgmental language.
- Policy 5: Indicate the relative prominence of
opposing views.

Note that even if it slightly violates the NPOV
policy, it is considered biased and not neutral.

EXAMPLES
{example_str}

EDIT
{edit_here}
\end{verbatim}
\end{scriptsize}

\subsection{Detection: NPOV Scoped}
\textit{Note: The EXAMPLES part of the prompt is only included for the few-shot conditions.} \\
\begin{scriptsize}
\begin{verbatim}
INSTRUCTIONS
Act like an expert Wikipedia editor. Classify if
this Wikipedia edit is neutral or biased.

A Wikipedia edit is biased if it violates
Wikipedia's neutral point of view (NPOV) policy.

Neutral point of view policy:
- Policy 1: Avoid subjective and judgmental
language. Word choice should use a disinterested,
impartial tone and avoid language that sympathizes
with or disparages the subject.
- Policy 2: Avoid phrasing that implies how
‘believable’ a statement is. Word choice can imply
how believable the author considers a statement to
be.
- Policy 3: Avoid stereotypes. Word choice can
imply stereotypes about gender, race or other
demographic categories.
- Policy 4: Avoid presenting opinions as facts.
Opinions should be attributed to a person, a group
of people who hold the opinion, or where
appropriate, described as a widespread view.

Note that even if it slightly violates the NPOV
policy, it is considered biased and not neutral.

EXAMPLES
{example_str}

EDIT
{edit_here}

RETURN
a json like {"response": "biased" or "neutral",
"policy_violated": "none" if neutral else policy
violated, "reasoning": reasoning} and nothing else
\end{verbatim}
\end{scriptsize}

\subsection{Detection: Minimal}
\textit{Note: The EXAMPLES part of the prompt is only included for the few-shot conditions}. \\
\begin{scriptsize}
\begin{verbatim}
INSTRUCTIONS
Act like an expert encyclopedia editor. Classify
if a given statement is neutral or biased.

Note that even if it slightly violates the NPOV
policy, it is considered biased and not neutral.

EXAMPLES
{example_str}

STATEMENT
{edit_here}

RETURN
a json like {"response": "biased" or "neutral",
"reasoning": reasoning} and nothing else
\end{verbatim}
\end{scriptsize}

\subsection{Detection: LLM Bootstrapped Prompt (NPOV-Scoped)}
\textit{Note: This prompt format follows DSPY's Chain of Thought implementation.}\\
\begin{scriptsize}
\begin{verbatim}
Act like an expert Wikipedia editor. Classify if
this Wikipedia edit is neutral or biased. A
Wikipedia edit is biased if it violates
Wikipedia's neutral point of view (NPOV) policy.

Neutral point of view policy:
- Policy 1: Avoid subjective and judgmental
language. Word choice should use a disinterested,
impartial tone and avoid language that sympathizes
with or disparages the subject.
- Policy 2: Avoid phrasing that implies how
'believable' a statement is. Word choice can imply
how believable the author considers a statement to
be.
- Policy 3: Avoid stereotypes. Word choice can
imply stereotypes about gender, race or other
demographic categories.
- Policy 4: Avoid presenting opinions as facts.
Opinions should be attributed to a person, a group
of people who hold the opinion, or where
appropriate, described as a widespread view. Note
that even if it slightly violates the NPOV policy,
it is considered biased and not neutral.

---

Follow the following format.

Edit: edit
Reasoning: Let's think step by step in order to
${produce the answer}. We ...
Answer: neutral or biased

---

Edit: hydrogen pipelines are unfortunately more
expensive than even long-distance electric lines.
Reasoning: Let's think step by step in order to
produce the answer. We first identify if the edit
uses subjective or judgmental language. The word
"unfortunately" introduces a subjective viewpoint,
implying a negative aspect of hydrogen pipelines
compared to electric lines. This word choice
violates Policy 1 by not maintaining an impartial
tone. It also subtly presents an opinion as a
fact, which is against Policy 4, as it does not
attribute this viewpoint to a specific source or
acknowledge it as a perspective rather than an
objective truth. Therefore, the edit does not
adhere to the neutral point of view policy.
Answer: biased

---

(...Omitting more examples for space...)

---

Edit: {EDIT_HERE} Reasoning: Let's think step by
step in order to
\end{verbatim}
\end{scriptsize}

\subsection{Detection: LLM Bootstrapped Prompt (NPOV)}
\textit{Note: This prompt format follows DSPY's Chain of Thought implementation.}\\
\begin{scriptsize}
\begin{verbatim}
Act like an expert Wikipedia editor. Classify if
this Wikipedia edit is neutral or biased. A
Wikipedia edit is biased if it violates
Wikipedia's neutral point of view (NPOV) policy.
Neutral point of view policy:
- Policy 1: Avoid stating opinions as facts
- Policy 2: Avoid stating seriously contested
assertions as facts
- Policy 3: Avoid stating facts as opinions
- Policy 4: Prefer nonjudgmental language.
- Policy 5: Indicate the relative prominence of
opposing views.
Note that even if it slightly violates the NPOV
policy, it is considered biased and not neutral.

---

Follow the following format.

Edit: edit
Reasoning: Let's think step by step in order to
${produce the answer}. We ...
Answer: neutral or biased

---

Edit: he developed a complex story-line with
theological themes, as adam warlock, who had
emerged from a kind of virgin birth, confronted a
dictatorial universal church of truth founded by
his evil future self.

Reasoning: Let's think step by step in order to
produce the answer. First, we examine if the edit
states opinions as facts. The edit provides a
summary of a storyline involving Adam Warlock, a
character from comic books, and describes the plot
and themes involved. It does not present opinions
as facts but rather summarizes the content of a
fictional work. Second, we check if it states
seriously contested assertions as facts. The edit
does not make assertions about real-world events
or figures but discusses a fictional narrative.
Third, we consider if it states facts as opinions.
The edit does not frame factual information as
opinion. Fourth, we evaluate the use of
nonjudgmental language. The language used to
describe the storyline and themes is neutral and
descriptive without passing judgment. Fifth, we
assess if it indicates the relative prominence of
opposing views. Since this edit is about a
specific storyline within a comic book, the
concept of opposing views is not applicable in the
same way it would be for a real-world issue.
Therefore, based on these considerations, the edit
does not violate the NPOV policy.

Answer: neutral

---


(...Omitting more examples for space...)

---

Edit: {EDIT_HERE}
Reasoning: Let's think step by step in order to
\end{verbatim}
\end{scriptsize}

\subsection{Detection: LLM Auto-Generated Prompt}
\textit{Note: This prompt format follows DSPY's Chain of Thought implementation.}\\

\begin{scriptsize}
\begin{verbatim}
Please meticulously evaluate the content of the
indicated Wikipedia edit. Our aim is to determine
its alignment with the principles of neutrality
and objective information presentation. An edit
falls under "neutral" if it maintains or improves
the essential impartial perspective of the topic
by offering well-referenced data, balanced
interpretation, and avoids espousing a specific
ideology, viewpoint, or favoritism towards any
subject matter or entity portrayed. Conversely,
categorize an edit as "biased" when it skews
towards advocating a singular perspective:
conspicuously highlighted by the use or omission
of particular terminologies, incorporation or
deletion of facts that possess agenda-driven
subtexts, or through nuanced alteration playing to
subjective opinions, ideologies, or non-verifiable
anecdotes that conspicuously colour the narrative
away from evidentiary neutrality. In making your
categorization, lavish detail on linguistic
nuances, presentation styles, and contextual frame
shifts evoked by the edit, analyzing its departure
from or congruity to non-partisan, factual,
information dissemination aimed at fostering an
informed readership.


---

Follow the following format.

Edit: edit
Reasoning: Let's think step by step in order to
${produce the answer}. We ...
After thorough inspection, the Wikipedia edit's
alignment with neutral-objective editing standards
is: neutral or biased

---

Edit: {EDIT_HERE}
Reasoning: Let's think step by step in order to
\end{verbatim}
\end{scriptsize}

\subsection{Detection: LLM Bias-Based Prompt}
\label{bias_based}
\begin{scriptsize}
\begin{verbatim}
INSTRUCTIONS
Act like an expert encyclopedia editor. Classify
if a given statement is neutral or biased.

TYPES OF BIAS
- Framing bias: Using subjective words or phrases
linked with a particular point of view (using
words like best or deepest or using pilfered from
instead of based on)

- Demographic bias: text with presuppositions
about particular genders, races, or other
demographic categories (like presupposing that all
programmers are male)

- Epistemological bias: linguistic features that
subtly (often via presupposition) modify the
believability of a proposition

Note that even if it is slightly biased, it is
considered biased and not neutral.

STATEMENT
{edit_here}

RETURN
Return only a valid JSON in the format:
{"response": "biased" or "neutral", "bias": type
of bias if biased else "none", "explanation":
one-sentence reason why you make the bias
decision}
\end{verbatim}
\end{scriptsize}

\subsection{Generation: NPOV Scoped}

\textit{Note: The CONSTRAINTS section of the prompt is only included in the `Is Constrained' condition.}

\begin{scriptsize}
    
\begin{verbatim}
Given a biased Wikipedia edit, make minimal
changes to make this edit neutral according to
Wikipedia's neutral point of view (NPOV) policy.

Neutral point of view policy:
- Policy 1: Avoid subjective and judgmental
language. Word choice should use a disinterested,
impartial tone and avoid language that sympathizes
with or disparages the subject.
- Policy 2: Avoid phrasing that implies how
'believable' a statement is. Word choice can imply
how believable the author considers a statement to
be.
- Policy 3: Avoid stereotypes. Word choice can
imply stereotypes about gender, race or other
demographic categories.
- Policy 4: Avoid presenting opinions as facts.
Opinions should be attributed to a person, a group
of people who hold the opinion, or where
appropriate, described as a widespread view.

CONSTRAINTS
- Change as few words as possible
- Make minimal changes

EDIT
{EDIT}

RETURN an extractable json like
{'neutral_edit':EDIT}
\end{verbatim}
\end{scriptsize}
\subsection{Generation: CAI}

\textit{Note: the CONSTRAINTS section of the prompt is only included in the `Is Constrained' condition.}

\subsubsection{Critique request}
\begin{scriptsize}
\begin{verbatim}
Identify specific ways, writing in one paragraph,
in which this Wikipedia edit violates Wikipedia's
neutral point of view (NPOV) policy.

Neutral point of view policy:
- Policy 1: Avoid subjective and judgmental
language. Word choice should use a disinterested,
impartial tone and avoid language that sympathizes
with or disparages the subject.
- Policy 2: Avoid phrasing that implies how
‘believable’ a statement is. Word choice can imply
how believable the author considers a statement to
be.
- Policy 3: Avoid stereotypes. Word choice can
imply stereotypes about gender, race or other
demographic categories.
- Policy 4: Avoid presenting opinions as facts.
Opinions should be attributed to a person, a group
of people who hold the opinion, or where
appropriate, described as a widespread view.

EDIT
{EDIT}

RETURN a json like {'critique':critique}
\end{verbatim}
\end{scriptsize}

\subsubsection{Revision request}
\begin{scriptsize}
\begin{verbatim}
Please rewrite the wikipedia edit based on this critique. 

CRITIQUE
{CRITIQUE}

EDIT
{EDIT}

CONSTRAINTS 
- Change as few words as possible
- Make minimal changes

RETURN a json like {'new_edit':new_edit}
\end{verbatim}
\end{scriptsize}

\clearpage

\section{LLM Bias Detection}
\begin{figure*}[h!]
    \centering
    \includegraphics[width=1\linewidth]{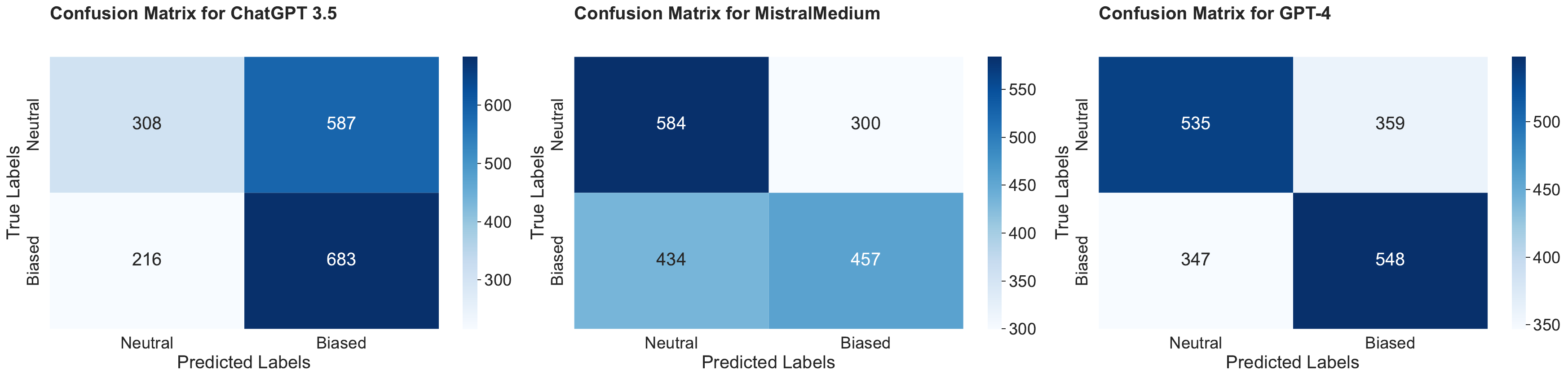}
    \caption{Confusion matrices from classification experiment.}
    \label{fig:confusion}
\end{figure*}

\begin{table}[h!]
\footnotesize
\centering
\begin{tabular}{lrrr}
\toprule
              Topic &  Total &  Biased &  Neutral \\
\midrule
            Culture &        56296 &         28148 &          28148 \\
          Geography &        28348 &         14174 &          14174 \\
History And Society &        20986 &         10493 &          10493 \\
               STEM &         8722 &          4361 &           4361 \\
         Assistance &          120 &            60 &             60 \\
\bottomrule
\end{tabular}
\caption{Number of edits for each topic. The data consists of rewrites of biased edits.}
\label{tab:topic_counts}
\end{table}

\subsection{LLM Self-Optimizations}
\label{self_opt}
We experimented with more advanced reasoning techniques and LLM self-optimizations (see Appendix \ref{appendix_prompts} for these LLM-assisted prompts). First, we selected our top model (GPT-4) and the NPOV and NPOV-Scoped prompts to use with chain-of-thought (CoT) reasoning and DSPy's `BootstrapFewShot' module. In chain-of-thought reasoning, we explicitly instruct the model to show its step-by-step thinking process (like showing work on a math problem) rather than just providing an answer. This approach often improves model performance~\cite{wei_chain--thought_2023}. DSPy~\cite{khattab_dspy_2023} is a state-of-the-art framework for self-improving prompt pipelines. In DSPy's BootstrapFewShot module, the LLM will learn rationales for labeled instances in a training phase. Then on the test set, instead of simply seeing an example and a label (standard few-shot), the LLM will also see these `bootstrapped' rationales when making classifications. For both prompts, we tested 300 edits with a 70-30 train-test split and 20-shot bootstrapped rationales. (We doubled the number of examples from our 10-shot experiment to amplify the effect of LLM-generated rationales.) The 20-shot augmented NPOV-Scoped prompt achieved 64\% accuracy on the held-out set (+1\% higher but statistically indistinguishable from the best unaugmented prompt); the 20-shot augmented NPOV-base prompt achieved 62\% accuracy. We also experimented with an LLM optimizing its prompt on its own. To do this, we used the COPRO module from DSPy~\cite{khattab_dspy_2023}. Briefly, a model is initialized with an initial minimal prompt, and then in a training phase the model repeatedly generates and refines the prompt based on performance on a subset of examples. We similarly used 300 examples with a 70-30 train-test split. The top AI-generated prompt from the training phase achieved a test-set accuracy of 61\%. For the rest of the detection analysis, we analyze the results from our original experiments.

\subsection{LLM Bias Identification Approach}

We also experimented with another prompting approach (Appendix \ref{bias_based} for prompt). The idea here was to first have the LLM identify the kind of bias (if there was one), and \textit{then} make a judgment. We used the biases and definitions from \cite{pryzant_automatically_2020}. Using the same set of edits and GPT-4 (our top-performing model), this approach yielded an accuracy of 0.60.

\subsection{Model Level Analysis}

\label{expl_level_app}
\begin{figure*}[h!]
    \centering
    \includegraphics[width=0.8\linewidth]{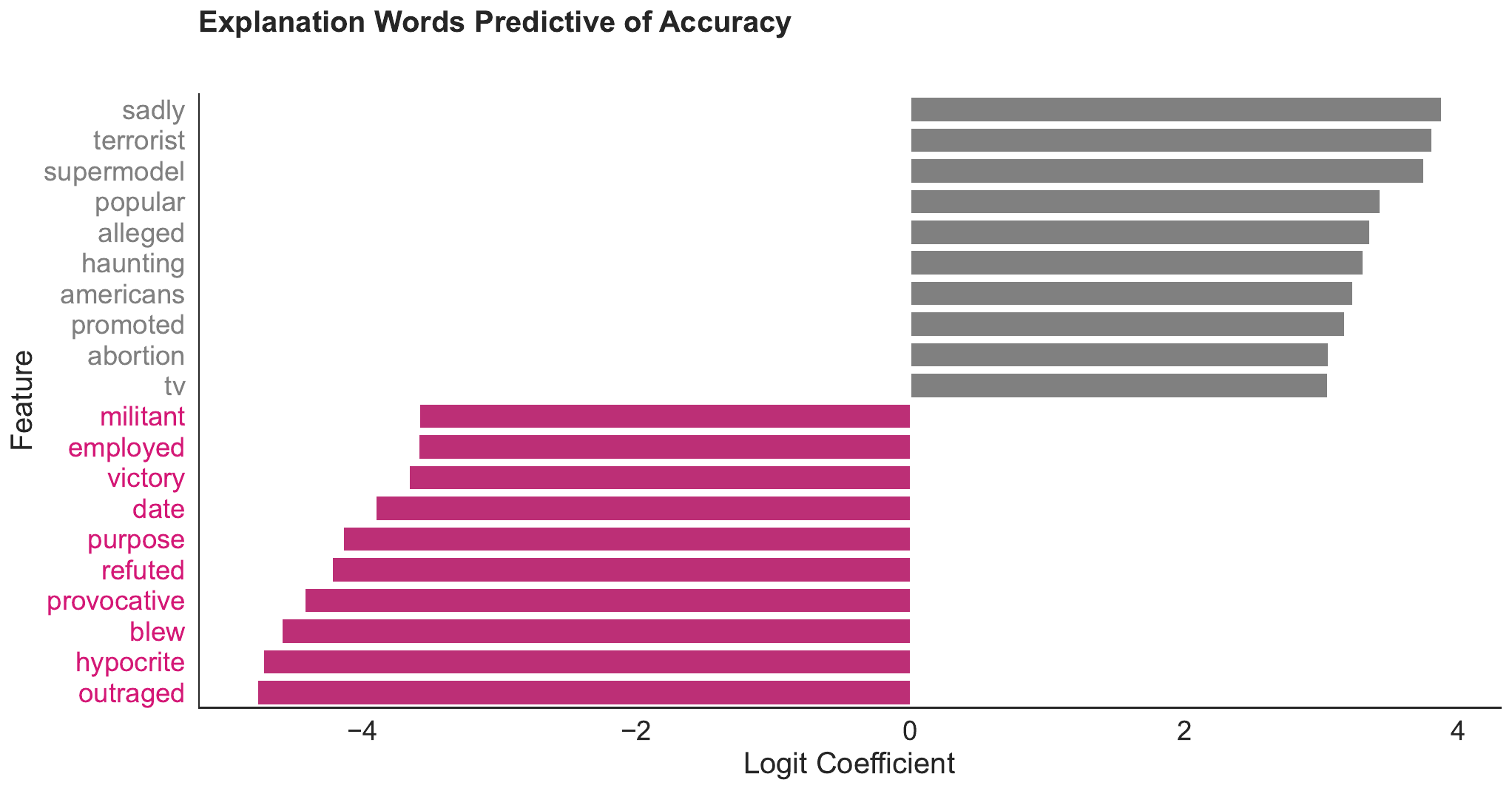}
    \caption{Words in an explanation with the most negative and most positive logit coefficients after a TF-IDF logistic regression predicting accuracy. Positive coefficients are associated with accuracy.}
    \label{fig:logit_explanation}
\end{figure*}

\paragraph{Tests for Model Error Imbalance.}
\label{binom_test}
We assessed the statistical significance of model error patterns. Assume a null hypothesis (HO) where a model's errors are equally likely to be false positives (FPs) or false negatives (FNs). (Here, an FP means falsely predicting that an edit is biased and an FN means falsely predicting that an edit is neutral.) Under this null hypothesis, the proportion of errors that are FPs should be 0.5. We tested whether a model's errors significantly deviated from this proportion using a two-tailed binomial test. In Figure \ref{fig:binom} we visualize the distribution of FPs under a null hypothesis that $FP=FN$ against the actual proportion of errors that are FPs. We reject the null hypothesis for ChatGPT 3.5 ($p<0.001$) and Mistral-Medium ($p<0.001$), but not GPT-4 ($p=0.68)$--- indicating that the first two models are more prone to specific types of errors. See Appendix Figure \ref{fig:confusion} for confusion matrices.

\subsection{Edit Level Analysis}
\label{edit_level_app}

We conducted a qualitative analysis of `easy' vs. `hard' biased edits (i.e., top or bottom PCC quartile). Top-PCC biased edits tend to have some highly subjective word that alerts models these edits are biased. See below (emphasis added). 

\begin{itemize}  
\item \textit{``one of the central characters of the novel, akili kuwale, provides a \textbf{brilliant} demonstration of this change and its implications, together with \textbf{excellent} characterization.''}
\item \textit{``environmentalists \textbf{complain} that before shipbreaking began there in june 1983 the beach at alang was pristine and unspoiled.''}
\item \textit{``colchester has a \textbf{proud} tradition of its citizen volunteers serving in the territorial army.''}
\end{itemize}

By contrast, low-PCC biased edits generally do not contain an overtly subjective adjective.

\begin{itemize} 
\item \textit{``the bill protects americans against discrimination based on their genetic information when it comes to health insurance and employment.''}
\item \textit{``confucianism (; pinyin: rxu ; literally means ``the school of the scholars'', see also names for confucianism for details) is an east asian ethical and philosophical system derived from the teachings of the early chinese sage confucius.''}
\item \textit{``they can see far over the great plains of illinois and across lake michigan on a clear day.''}

\end{itemize}

\begin{figure*}[h!]
    \centering
    \includegraphics[width=0.9\linewidth]{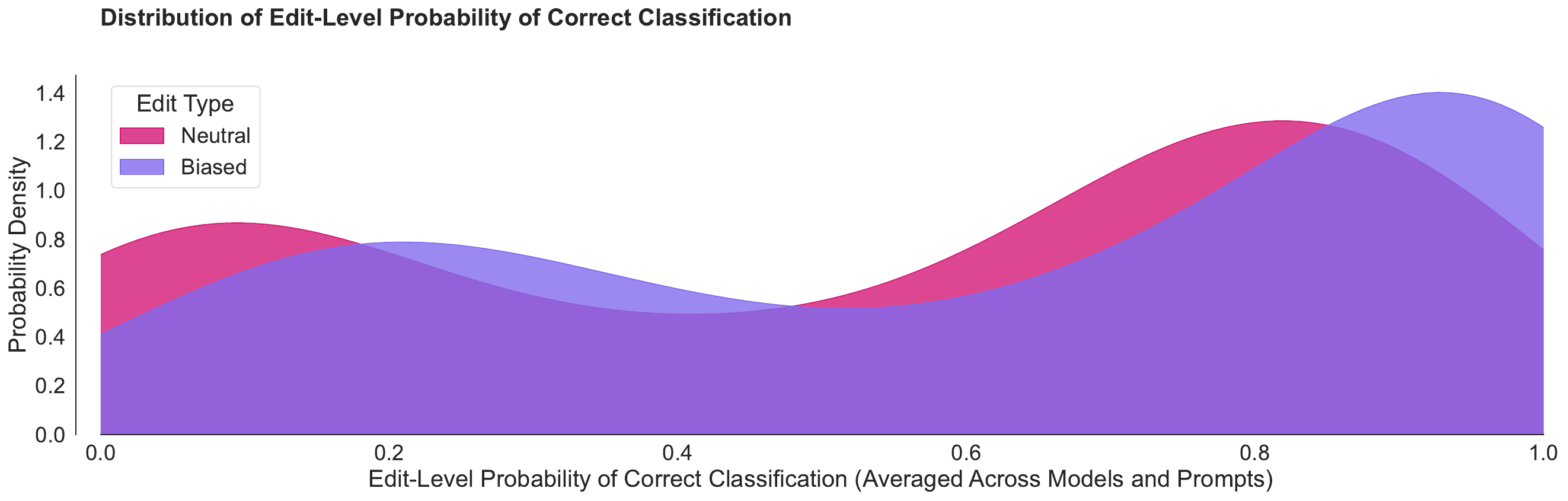}
    \caption{Edit difficulty was bimodal and models were more accurate for biased edits.}
    \label{ridge}
\end{figure*}

\subsection{Explanation Level Analysis: Quantitative Analysis} 
\subsubsection{TF-IDF Regression}
To understand what words in model explanations are correlated with prediction accuracy, we conducted a TF-IDF regression (Section 3). Important features are displayed in Appendix Figure \ref{fig:logit_explanation}.

\begin{table*}[!htbp] \centering
\begin{tabular}{@{\extracolsep{5pt}}lc}
\\[-1.8ex]\hline
\hline \\[-1.8ex]
& \multicolumn{1}{c}{\textit{Dependent variable: Accuracy}} \
\cr \cline{2-2}
\\[-1.8ex] & (1) \\
\hline \\[-1.8ex]
 EditDistance[high] & 1.433$^{***}$ \\
& (1.211 , 1.696) \\
 EditDistance[med] & 1.863$^{***}$ \\
& (1.605 , 2.162) \\
 Examples[FewShot] & 1.005$^{}$ \\
& (0.901 , 1.122) \\
 Intercept & 0.795$^{**}$ \\
& (0.651 , 0.971) \\
 Model[GPT-4] & 1.248$^{***}$ \\
& (1.091 , 1.427) \\
 Model[MistralMedium] & 1.150$^{**}$ \\
& (1.006 , 1.315) \\
 Prompt[NPOVScoped] & 0.974$^{}$ \\
& (0.851 , 1.114) \\
 Prompt[NPOV] & 0.962$^{}$ \\
& (0.842 , 1.101) \\
 RootTopic[Culture] & 1.447$^{***}$ \\
& (1.215 , 1.724) \\
 RootTopic[Geography] & 1.011$^{}$ \\
& (0.850 , 1.203) \\
 RootTopic[HistoryAndSociety] & 1.074$^{}$ \\
& (0.903 , 1.277) \\
 RootTopic[STEM] & 1.314$^{***}$ \\
& (1.103 , 1.565) \\
 ZScoreEditWordCount & 1.098$^{***}$ \\
& (1.023 , 1.177) \\
\hline \\[-1.8ex]
 Observations & 5358 \\
 Pseudo $R^2$ & 0.016 \\
\hline
\hline \\[-1.8ex]
\textit{Note:} & \multicolumn{1}{r}{ $^{**}$p$<$0.05; $^{***}$p$<$0.01} \\
\end{tabular}
\caption{Logistic regression predicting accuracy. The reference level for edit distance is `low', the reference model is ChatGPT 3.5, the reference prompt is `minimal', FewShot is compared to ZeroShot, and the reference topic is `Assistance'. The word count of the edit is z-scored. ORs are shown with 95\% CIs are in parentheses.}
\label{logitacc}
\end{table*}

\pagebreak

\subsection{Explanation Level Analysis: Qualitative Analysis} 
\label{qual_exp}
A general pattern we found is that LLMs tended to misclassify edits as biased when there was a highly subjective phrase, even if that phrase was being ascribed to another entity. Apart from the example in the main manuscript (about an ad calling Obama a hypocrite), consider this edit:

\begin{quote}
an example of anti-necktie sentiment is found in iran, whose theocratic rulers have denounced the accessory as a decadent symbol of western oppression.
\end{quote}

Here is Mistral-Medium's explanation: 

\begin{quote}
The edit uses judgmental language to describe neckties as a 'decadent symbol of western oppression'. This is a value judgment that presents a particular point of view, rather than a neutral description of the facts.
\end{quote}

But the edit is \textit{not} describing neckties as `a decadent symbol of Western oppression', rather relaying what a regime thinks of them. 

Another interesting example is where an edit is not necessarily violating NPOV, but it contains words that are likely to be associated with controversy. Here is an example edit:

\begin{quote}
lehi (group), zionist paramilitary organization in british mandate of palestine
\end{quote}

GPT-4 correctly identifies this edit as neutral:
\begin{quote}
    The edit simply states factual information without implying any judgment or opinion, adhering to a neutral point of view.
\end{quote}

But we see ChatGPT thinks that because the edit contains ``zionist'', it is NPOV-violating---even though a source such as Britannca would also classify the Lehi group identifies as Zionist\footnote{\url{https://www.britannica.com/topic/Stern-Gang}}.

\begin{quote}
     The term 'Zionist paramilitary organization' is a seriously contested assertion as it can be seen as promoting one particular viewpoint on the organization's activities and goals. It would be more neutral to simply state 'Lehi, a paramilitary organization in British Mandate of Palestine.'
\end{quote}

However, LLMs were generally quite accurate at flagging egregiously NPOV-violating terms. Here is an example of an edit:

\begin{quote}
one of the central characters of the novel, akili kuwale, provides a brilliant demonstration of this change and its implications, together with excellent characterization
\end{quote}

Here is GPT-4's explanation:

\begin{quote}
    Using words like 'brilliant' and 'excellent' to describe the demonstration and characterization injects a subjective judgment, violating the nonjudgmental language requirement
\end{quote}

Here is Mistral-Medium's explanation:

\begin{quote}
    The edit uses judgmental language by describing Akili Kuwale's demonstration as 'brilliant' and the characterization as 'excellent'. This violates the neutral point of view policy by expressing a positive opinion as fact
\end{quote}

\onecolumn

\noindent
\begin{minipage}[t]{0.45\textwidth}
    \section{LLM Neutrality Generation: Computational Evaluation}
\end{minipage}

\begin{table}[h!] 
\centering
\scriptsize
\begin{tabular}{@{\extracolsep{5pt}}lccc}
\\[-1.8ex]\hline
\hline \\[-1.8ex]
\\[-1.8ex] & \multicolumn{1}{c}{Edit Distance} & \multicolumn{1}{c}{Word Count} & \multicolumn{1}{c}{N Changes}  \\
\\[-1.8ex] & (1) & (2) & (3) \\
\hline \\[-1.8ex]
 Intercept & 0.556$^{***}$ & 47.270$^{***}$ & 25.880$^{***}$ \\
& (0.535 , 0.577) & (44.539 , 50.001) & (24.404 , 27.356) \\
 IsConstrained[Y] & -0.209$^{***}$ & -17.100$^{***}$ & -12.940$^{***}$ \\
& (-0.239 , -0.179) & (-20.962 , -13.238) & (-15.028 , -10.852) \\
 IsConstrained[Y]:PromptType[ZeroShot] & 0.102$^{***}$ & 16.500$^{***}$ & 8.610$^{***}$ \\
& (0.059 , 0.144) & (11.039 , 21.961) & (5.657 , 11.563) \\
 PromptType[ZeroShot] & -0.270$^{***}$ & -18.620$^{***}$ & -15.365$^{***}$ \\
& (-0.300 , -0.240) & (-22.482 , -14.758) & (-17.453 , -13.277) \\
\hline \\[-1.8ex]
 Observations & 800 & 800 & 800 \\
 $R^2$ & 0.447 & 0.142 & 0.324 \\
 Adjusted $R^2$ & 0.445 & 0.138 & 0.321 \\
 Residual Std. Error & 0.153 (df=796) & 19.673 (df=796) & 10.637 (df=796) \\
 F Statistic & 214.783$^{***}$ (df=3; 796) & 43.738$^{***}$ (df=3; 796) & 126.924$^{***}$ (df=3; 796) \\
\hline
\hline \\[-1.8ex]
\textit{Note:} & \multicolumn{3}{r}{ $^{**}$p$<$0.05; $^{***}$p$<$0.01} \\
\end{tabular}
\caption{OLS regressions of intensity metrics. 95\% CIs in parentheses.}
\label{regautomet}
\end{table}

\begin{table*}[!h] \centering
\tiny
\setlength{\tabcolsep}{2pt} 
\begin{tabular}{@{\extracolsep{5pt}}lccccc}
\\[-1.8ex]\hline
\hline \\[-1.2ex]
\\[-1.8ex] & \multicolumn{1}{c}{P} & \multicolumn{1}{c}{R} & \multicolumn{1}{c}{S} & \multicolumn{1}{c}{N} & \multicolumn{1}{c}{B}  \\
\\[-1.8ex] & (1) & (2) & (3) & (4) & (5) \\
\hline \\[-1.8ex]
 Intercept & 0.289$^{***}$ & 0.886$^{***}$ & 0.275$^{***}$ & 0.900$^{***}$ & 0.230$^{***}$ \\
& (0.246 , 0.332) & (0.837 , 0.934) & (0.234 , 0.315) & (0.849 , 0.951) & (0.200 , 0.260) \\
 IsConstrained[Y] & 0.070$^{**}$ & -0.070$^{**}$ & 0.053$^{*}$ & -0.040$^{}$ & 0.213$^{***}$ \\
& (0.009 , 0.131) & (-0.139 , -0.001) & (-0.004 , 0.110) & (-0.112 , 0.032) & (0.170 , 0.255) \\
 IsConstrained[Y]:PromptType[ZeroShot] & 0.029$^{}$ & -0.031$^{}$ & 0.015$^{}$ & -0.025$^{}$ & -0.059$^{*}$ \\
& (-0.057 , 0.115) & (-0.129 , 0.066) & (-0.066 , 0.096) & (-0.127 , 0.077) & (-0.118 , 0.001) \\
 PromptType[ZeroShot] & 0.085$^{***}$ & -0.113$^{***}$ & 0.070$^{**}$ & -0.075$^{**}$ & 0.253$^{***}$ \\
& (0.024 , 0.146) & (-0.182 , -0.044) & (0.012 , 0.127) & (-0.147 , -0.003) & (0.210 , 0.295) \\
\hline \\[-1.8ex]
 Observations & 800 & 752 & 796 & 800 & 800 \\
 $R^2$ & 0.043 & 0.050 & 0.028 & 0.019 & 0.313 \\
 Adjusted $R^2$ & 0.040 & 0.046 & 0.024 & 0.016 & 0.311 \\
 Residual Std. Error & 0.310 (df=796) & 0.340 (df=748) & 0.291 (df=792) & 0.367 (df=796) & 0.216 (df=796) \\
 F Statistic & 11.972$^{***}$ (df=3; 796) & 13.033$^{***}$ (df=3; 748) & 7.567$^{***}$ (df=3; 792) & 5.220$^{***}$ (df=3; 796) & 121.098$^{***}$ (df=3; 796) \\
\hline
\hline \\[-1.8ex]
\textit{Note:} & \multicolumn{5}{r}{$^{*}$p$<$0.1; $^{**}$p$<$0.05; $^{***}$p$<$0.01} \\
\end{tabular}
\caption{OLS regressions of edit-level accuracy NLG metrics, which are computed based on comparing the sets of words that AI and humans removed from edits. P,R,S,N,B are precision, recall, similarity, non-disjoint, and BLEU score. 95\% CIs in parentheses.}
\label{regautometacc}
\end{table*}
\begin{table*}[!htbp] \centering
\tiny
\begin{tabular}{@{\extracolsep{5pt}}lcc}
\\[-1.8ex]\hline
\hline \\[-1.8ex]
\\[-1.8ex] & \multicolumn{1}{c}{Count of Added Words} & \multicolumn{1}{c}{Count of Removed Words}  \\
\\[-1.8ex] & (1) & (2) \\
\hline \\[-1.8ex]
 Intercept & 17.455$^{***}$ & 8.425$^{***}$ \\
& (16.421 , 18.489) & (7.834 , 9.016) \\
 IsConstrained[Y] & -10.630$^{***}$ & -2.310$^{***}$ \\
& (-12.093 , -9.167) & (-3.146 , -1.474) \\
 IsConstrained[Y]:PromptType[ZeroShot] & 8.250$^{***}$ & 0.360$^{}$ \\
& (6.181 , 10.319) & (-0.822 , 1.542) \\
 PromptType[ZeroShot] & -12.315$^{***}$ & -3.050$^{***}$ \\
& (-13.778 , -10.852) & (-3.886 , -2.214) \\
\hline \\[-1.8ex]
 Observations & 800 & 800 \\
 $R^2$ & 0.364 & 0.151 \\
 Adjusted $R^2$ & 0.361 & 0.148 \\
 Residual Std. Error & 7.452 (df=796) & 4.257 (df=796) \\
 F Statistic & 151.733$^{***}$ (df=3; 796) & 47.118$^{***}$ (df=3; 796) \\
\hline
\hline \\[-1.8ex]
\textit{Note:} & \multicolumn{2}{r}{$^{*}$p$<$0.1; $^{**}$p$<$0.05; $^{***}$p$<$0.01} \\
\end{tabular}
\caption{OLS regressions of additions and removals, excluding stopwords. 95\% CIs in parentheses.}
\label{regautoaddrem}
\end{table*}

\begin{figure*}[h!]
    \centering
    \includegraphics[width=0.9\linewidth]{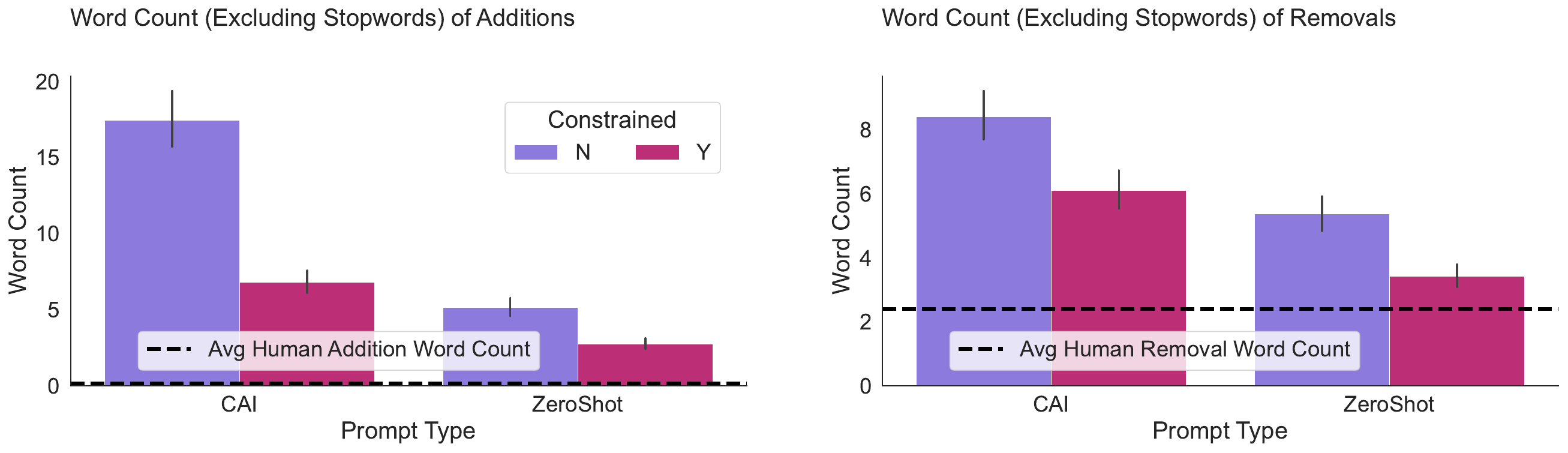}

    \caption{AI had both more removals and more additions than human editors. Error bars are 95\% CIs.}
    \label{fig:combined_change_plots_horizontal}
\end{figure*}

\begin{figure*}
    \centering
    \includegraphics[width=1\linewidth]{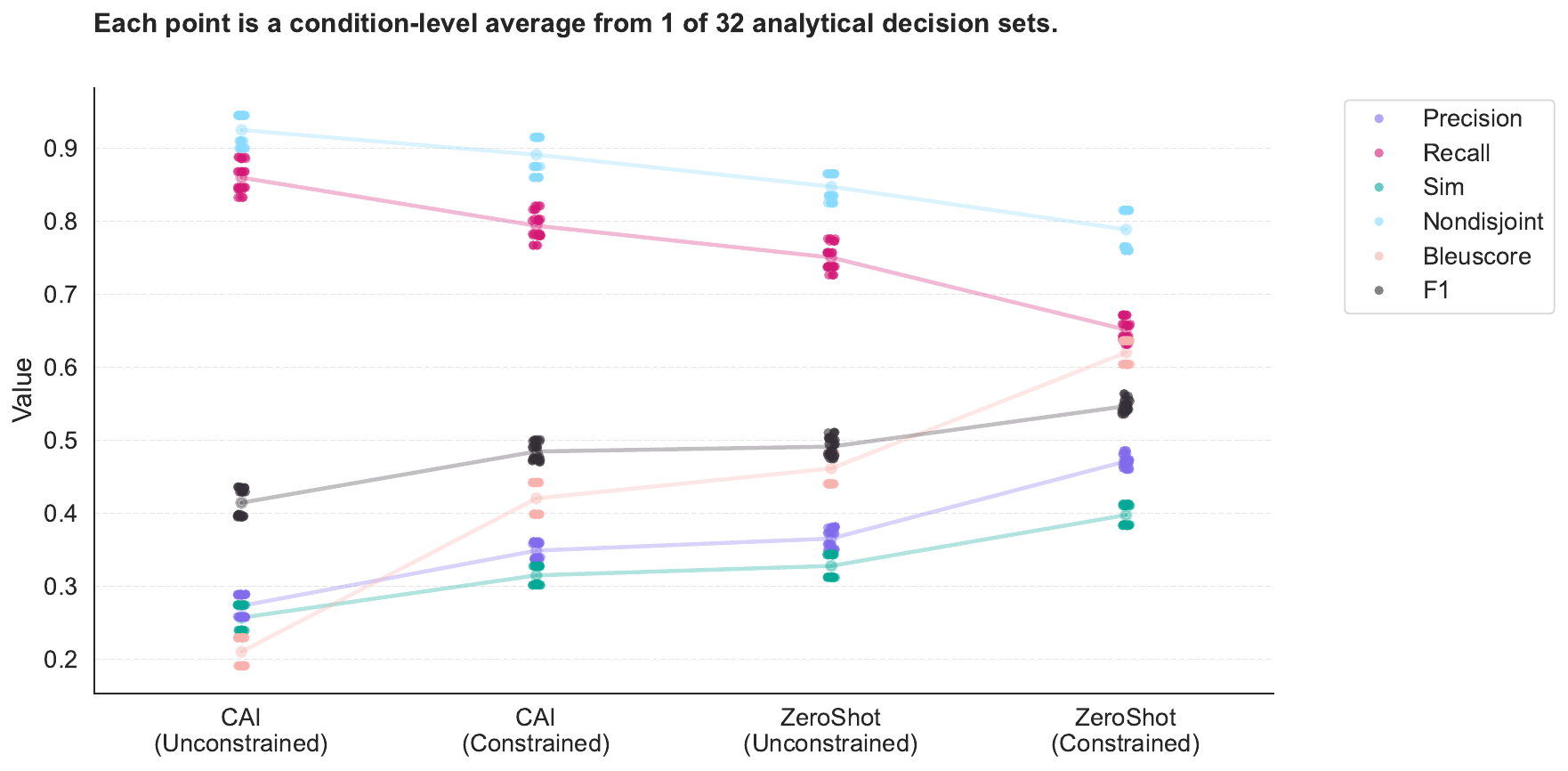}
    \caption{To rule out that our generation findings were dependent on seemingly minor analytical choices we made (such as stop-word removal), we re-ran this analysis 32 times under different analytic decisions. We varied (1) whether undefined precision calculations were treated as zero or missing, (2) whether undefined recall calculations were treated as zero or missing, (3) whether or not diff-based metrics stripped punctuation, (4) whether or not diff-based metrics stripped stop-words, and (5) whether BLEU scores were calculated with smoothing. Condition-level metrics (colors) are broadly similar regardless of analytical decision set. Each point is a condition-level average from one decision set.}
    \label{fig:multiverse}
\end{figure*}

\subsection{BERTScore}
\label{BertScore}
For our main analysis, we used a diff-based approach to test how LLMs vs humans edit differently, treating human changes as the gold standard. The benefit of a diff-based approach is that it is highly interpretable. The diff-based approach captures raw information changes---additions and  removals. However, it does not capture semantic change. Another way to analyze LLM changes is to compare their semantic content to Wikipedian changes. 

As a complementary analysis, we used BERTScore---a method that calculates precision and recall-type metrics via semantic similarity. From \cite{zhang2020bertscoreevaluatingtextgeneration}: For a reference $x$ and candidate $\hat{x}$, BERTScore score ``matches each token in $x$ to a token in $\hat{x}$ to compute recall, and each token in $\hat{x}$ to a token in $x$ to compute precision''. For a reference $x$ and candidate $\hat{x}$, recall and precision are:

\begin{align}
R_{\text{BERT}} = \frac{1}{|x|} \sum_{x_i \in x} \max_{x_j \in \hat{x}} \mathbf{x}_i^{\top} \mathbf{x}_j \, , \quad
P_{\text{BERT}} = \frac{1}{|\hat{x}|} \sum_{x_j \in \hat{x}} \max_{x_i \in x} \mathbf{x}_i^{\top} \mathbf{x}_j \, , \quad
\end{align}

where BERT is used to create embeddings. Here, we take the AI modified edit as the candidate and the human modified edit as the reference text. Using this approach, we directionally replicate our core finding: For AI edits, BertScore recall ($M$ = 0.94, $SD$ = 0.04) is higher than BertScore precision ($M$ = 0.93, $SD$ = 0.05), $t$(799) = 12.92, $p<0.001$, Cohen's $d$ = 0.46. But we note that relative to the pure diff-based approach, BERTScore yields directionally similar but smaller effects for AI precision vs recall. This is consistent with our human subject findings that AI generally does not change the semantic content of edits; AI editors just make many semantic-less changes. We discuss this in more detail in Discussion Takeaway 5.

\clearpage

\begin{minipage}[t]{0.45\textwidth}
    \section{LLM Neutrality Generation: Human Evaluation}

\end{minipage}

\begin{table*}[!htb]
\centering
\label{expr_demos_table}

\caption{Participant Demographics}
\begin{tabular}{lr}
\toprule
Characteristic & Percentage (\%) \\
\midrule
\multicolumn{2}{l}{\textit{Gender}} \\
\midrule
Woman & 57.1 \\
Man & 39.5 \\
Non-binary & 1.4 \\
Other & 1.4 \\
Prefer not to disclose & 0.7 \\
\midrule
\multicolumn{2}{l}{\textit{Age Range}} \\
\midrule
25-34 & 32.7 \\
35-44 & 26.5 \\
45-54 & 17.0 \\
18-24 & 12.9 \\
55-64 & 6.8 \\
65 or over & 4.1 \\
\midrule
\multicolumn{2}{l}{\textit{Educational Attainment}} \\
\midrule
Bachelor degree & 42.2 \\
Some college but no degree & 22.4 \\
High school degree or equivalent (e.g., GED) & 12.2 \\
Graduate degree (e.g., Masters, PhD, M.D) & 12.2 \\
Associate degree & 10.2 \\
Less than high school degree & 0.7 \\
\bottomrule
\end{tabular}
\end{table*}
\begin{table}[h] \centering 
\begin{tabular}{@{\extracolsep{5pt}}lcc} 
\\[-1.8ex]\hline 
\hline \\[-1.8ex] 
 & \multicolumn{2}{c}{\textit{Dependent variable:}} \\ 
\cline{2-3} 
\\[-1.8ex] & \multicolumn{2}{c}{} \\ 
 & Neutrality & Fluency \\ 
\\[-1.8ex] & (1) & (2)\\ 
\hline \\[-1.8ex] 
 conditionZeroShot & 0.854$^{***}$ & 0.613$^{***}$ \\ 
  & (0.080) & (0.077) \\ 
  & & \\ 
 conditionCAI & 0.802$^{***}$ & 0.276$^{***}$ \\ 
  & (0.080) & (0.075) \\ 
  & & \\ 
\hline \\[-1.8ex] 
Observations & 2,940 & 2,940 \\ 
R$^{2}$ & 0.076 & 0.027 \\ 
Max. Possible R$^{2}$ & 0.500 & 0.500 \\ 
Log Likelihood & $-$902.892 & $-$978.853 \\ 
Wald Test (df = 2) & 213.350$^{***}$ & 77.010$^{***}$ \\ 
LR Test (df = 2) & 232.068$^{***}$ & 80.146$^{***}$ \\ 
Score (Logrank) Test (df = 2) & 225.874$^{***}$ & 79.104$^{***}$ \\ 
\hline 
\hline \\[-1.8ex] 
\textit{Note:}  & \multicolumn{2}{r}{$^{*}$p$<$0.1; $^{**}$p$<$0.05; $^{***}$p$<$0.01} \\ 
\end{tabular} 
\caption{Conditional logistic regression models of the odds of choosing an AI edit over the human edit, for both neutrality and fluency. Results are on the log odds scale, with SEs in parentheses.} 
  \label{con_neut_flu} 
\end{table}

\begin{table*}
\centering
\begin{tabular}{lllll}
\toprule
Question Type &        Comparison & AI Proportion & Human Proportion &       Delta \\
\midrule
   Neutrality & Human vs ZeroShot &    0.70 &       0.30 &    0.40**** \\
   Neutrality &      Human vs CAI &    0.69 &       0.31 &    0.38**** \\
      Fluency & Human vs ZeroShot &    0.65 &       0.35 &    0.30**** \\
      Fluency &      Human vs CAI &    0.57 &       0.43 &     0.14*** \\
          Add & Human vs ZeroShot &    0.27 &       0.23 & 0.03 (n.s.) \\
          Add &      Human vs CAI &    0.34 &       0.33 & 0.01 (n.s.) \\
       Remove & Human vs ZeroShot &    0.41 &       0.40 & 0.01 (n.s.) \\
       Remove &      Human vs CAI &    0.42 &       0.40 & 0.02 (n.s.) \\
\bottomrule
\end{tabular}
\caption{Experiment results. For fluency and neutrality, participants chose between an AI and a human edit. P-values are computed using two-tailed binomial tests for whether the probability of picking an AI edit differs from 0.5. For additions and removals, participants evaluated each of the AI and huamn edits separately but were shown both at the same time. This table reports human addition and removal judgements at the matchup level. P-values are computed using chi-squared tests on whether human vs AI edits differed in frequencies of adding or removing information. `Delta' is the AI proportion minus the human proportion. Stars: n.s. $p>0.05, *p<0.05, **p<0.01 ***p<0.001, ****p<0.0001$}
\label{tab:sum_expr_stats_breakoutTrue}
\end{table*}

\begin{table*}
\centering
\begin{tabular}{llll}
\toprule
       Comparison &     Choice & Proportion Chosen (95\% CI) & Risk Ratio (95\% CI) \\
\midrule
     Human vs CAI &        CAI &    0.61 ([0.53, 0.69]) &  2.49 ([1.88, 3.48]) \\
     Human vs CAI &      Human &    0.25 ([0.18, 0.31]) &                  --- \\
     Human vs CAI & Both Equal &    0.15 ([0.09, 0.20]) &  0.59 ([0.35, 0.94]) \\
Human vs ZeroShot &   ZeroShot &    0.56 ([0.48, 0.64]) &  3.65 ([2.54, 5.85]) \\
Human vs ZeroShot &      Human &    0.15 ([0.10, 0.21]) &                  --- \\
Human vs ZeroShot & Both Equal &    0.28 ([0.21, 0.35]) &  1.83 ([1.18, 3.00]) \\
\bottomrule
\end{tabular}
\caption{Neutrality results from a `both-equal' pilot. Following our main experiment, we ran a pilot study (n=20; 15 trials each) where we used the same setup but gave participants an option to select `No substantial difference'. We only ran this pilot study for neutrality. The risk ratio is the ratio of the proportion of times an answer was chosen relative to the proportion of times the human rewrite was chosen.}
\label{neut_pilot_nochoice}
\end{table*}

\begin{table*}[t!]
\centering
\begin{tabular}{llllll}
\toprule
                   Subset & Question Type &        Comparison & \makecell{AI\\ Prop.} & \makecell{Human\\Prop.} &    Delta \\
\midrule
                Full data &       Fluency &      Human vs CAI &    0.57 &       0.43 &  0.14*** \\
           3 of 3 Correct &       Fluency &      Human vs CAI &    0.64 &       0.36 & 0.27**** \\
$>$25th Percentile Duration &       Fluency &      Human vs CAI &    0.55 &       0.45 &    0.11* \\
                Full data &       Fluency & Human vs ZeroShot &    0.65 &       0.35 & 0.30**** \\
           3 of 3 Correct &       Fluency & Human vs ZeroShot &    0.67 &       0.33 & 0.34**** \\
$>$25th Percentile Duration &       Fluency & Human vs ZeroShot &    0.64 &       0.36 & 0.29**** \\
                Full data &    Neutrality &      Human vs CAI &    0.69 &       0.31 & 0.38**** \\
           3 of 3 Correct &    Neutrality &      Human vs CAI &    0.75 &       0.25 & 0.50**** \\
$>$25th Percentile Duration &    Neutrality &      Human vs CAI &    0.68 &       0.32 & 0.37**** \\
                Full data &    Neutrality & Human vs ZeroShot &    0.70 &       0.30 & 0.40**** \\
           3 of 3 Correct &    Neutrality & Human vs ZeroShot &    0.75 &       0.25 & 0.49**** \\
$>$25th Percentile Duration &    Neutrality & Human vs ZeroShot &    0.70 &       0.30 & 0.39**** \\
\bottomrule
\end{tabular}
\caption{Robustness checks of main results. `Human Prop.' and `AI Prop.' denote the proportion of times the human or AI rewrite was chosen. P-values are computed using two-tailed binomial tests for whether the probability of picking an AI edit differs from 0.5. `Delta' is the AI proportion minus the human proportion. Subsets: All data; Participants who were above the 25th percentile in study duration; Participants who got all 3 training module questions correct (i.e., correctly classified edits as neutral or non-neutral). Stars: n.s. $p>0.05, *p<0.05, **p<0.01 ***p<0.001, ****p<0.0001$}
\label{tab:sum_expr_stats_rob_check}
\end{table*}

\begin{figure*}[h!]
    \centering
    \includegraphics[width=0.7\textwidth]{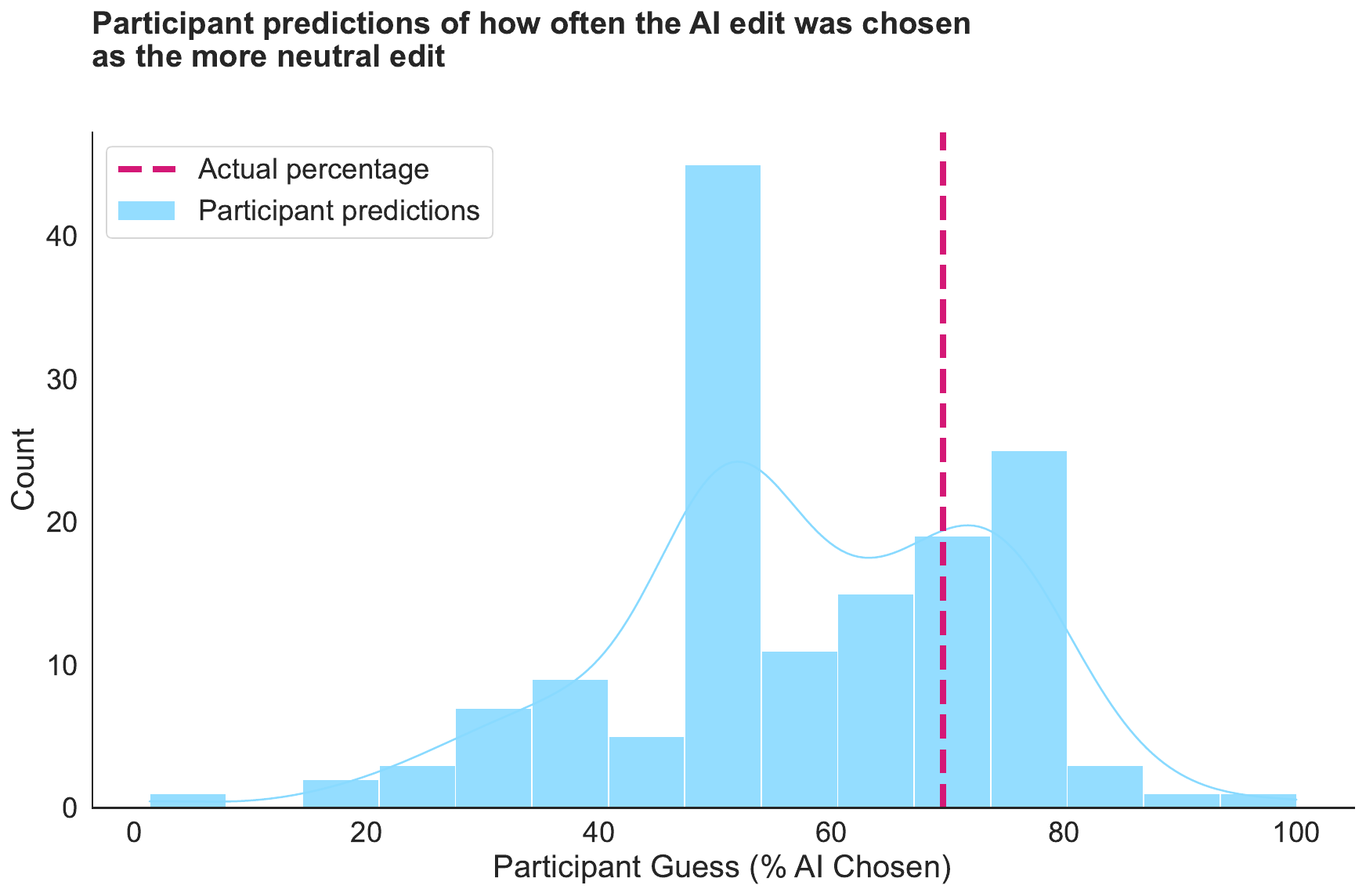}
    \caption{At the end of the experiment, participants guessed (via 0-100 slider) how often others chose the AI edit as more neutral than the human edit. The question was: ``For each pair of rewrites you saw, one rewrite was made by a Wikipedia editor and the other rewrite was made by a large language model (LLM) such as ChatGPT. Like you, other participants were not told which was which. What percent of the time do you think participants said the LLM rewrite increased neutrality more than the Wikipedia-editor rewrite?'' }
    \label{fig:neutrality_preds}
\end{figure*}

\end{document}